\documentclass{article}
\usepackage{neurips_2025}
\usepackage{graphicx} 
\usepackage{multirow} 
\usepackage{amssymb} 
\usepackage{siunitx}
\usepackage[utf8]{inputenc} 
\usepackage[T1]{fontenc}    
\usepackage{url}            
\usepackage{booktabs}       
\usepackage{amsfonts}       
\usepackage{nicefrac}       
\usepackage{microtype}      
\usepackage{xcolor}         
\usepackage{amssymb}
\usepackage{xcolor}
\usepackage{wrapfig}
\usepackage{subcaption}
\usepackage{tabularx}
\usepackage{enumitem}
\usepackage{float}
\usepackage{caption}
\usepackage{stfloats}
\usepackage{amsmath}
\usepackage{pifont}
\newcommand{\xmark}{\ding{55}} 
\newcommand{\cmark}{\ding{51}}
\usepackage[utf8]{inputenc}
\usepackage{hyperref}
\newcommand{\ours}[0]{K12Vista}
\definecolor{bb}{rgb}{0.12,0.565,1}
\definecolor{gg}{rgb}{0.2,0.8,0.2}
\definecolor{rr}{rgb}{1,0.85,0.2}

\title{\ours: Exploring the Boundaries of MLLMs in K-12 Education}
\author{
    \textbf{Chong Li}$^{1,2}$\thanks{Equal contribution. lichong@stu.pku.edu.cn, zhuchenglin@stu.pku.edu.cn, zhangtao.tanh@gmail.com.},
    \textbf{Chenglin Zhu}$^{1,2*}$,
    \textbf{Tao Zhang}$^{1*}$,
    \textbf{Mingan Lin}$^{1}$\thanks{Corresponding author. Correspondence to zenanchow@gmail.com, linmingan@baichuan-inc.com.},
    \textbf{Zenan Zhou}$^{1\dagger}$\\
    \small{$^{1}$Baichuan Inc, $^{2}$Peking University}
}

\begin{document}
\maketitle
\begin{abstract}
Multimodal large language models (MLLMs) have demonstrated remarkable reasoning capabilities in various visual tasks. However, their abilities in K12 (Grades 1–12) scenarios are still systematically underexplored. 
Previous studies suffer from various limitations including narrow subject coverage, insufficient data scale, lack of diversity in question types, and naive answer-centric evaluation method, resulting in insufficient exploration of model capabilities. 
To address these gaps, we propose \textbf{K12Vista}, the most comprehensive multimodal benchmark for Chinese K12 subject knowledge understanding and reasoning to date, featuring 33,000 questions across five core subjects from primary to high school and three question types. 
Moreover, beyond the final outcome, we are also concerned with the correctness of MLLMs' reasoning processes.
For this purpose, we meticulously compiles errors from MLLMs' reasoning processes and leverage an automated data pipeline to construct \textbf{K12-PEM-800K}, the largest process evaluation dataset offering detailed step-by-step judgement annotations for MLLMs' reasoning.
Subsequently, we developed \textbf{K12-PEM}, an advanced process evaluation model that integrates an overall assessment of both the reasoning process and answer correctness.
Moreover, we also introduce \textbf{K12-PEBench}, the first high-quality, human-annotated benchmark specifically designed for evaluating abilities of reasoning process evaluation.
Extensive experiments reveal that current MLLMs exhibit significant flaws when reasoning within K12Vista, providing critical insights for the development of more capable MLLMs. We open our resources at \url{https://github.com/lichongod/K12Vista}.
\end{abstract}
\section{Introduction}
\begin{figure*}[!t]
    \centering
    \includegraphics[width=0.95\textwidth]{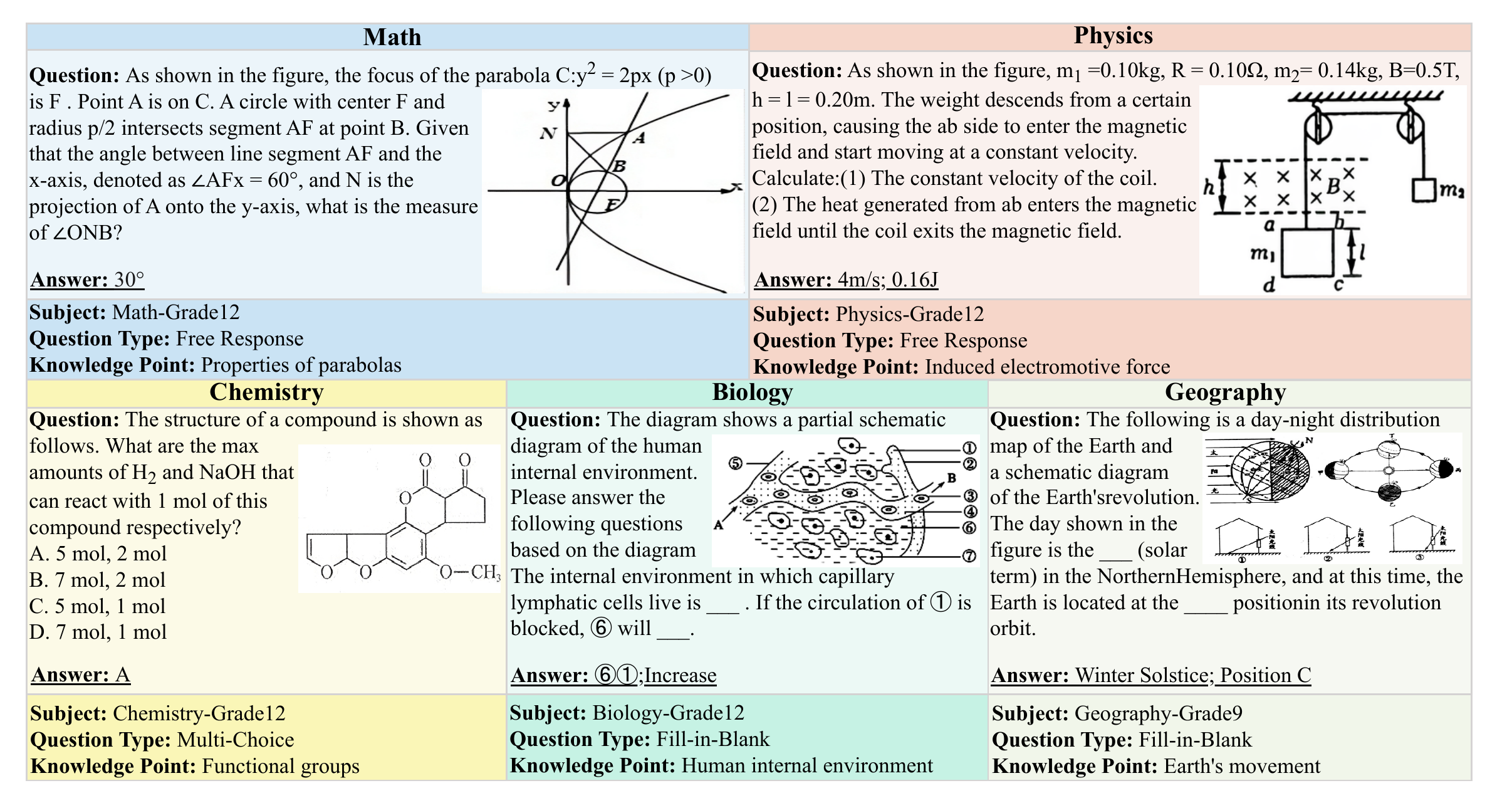}
    \caption{Some examples in K12Vista. Each question in K12Vista features high-quality text and images, offers diverse question types, and is enriched with attributes like subject and knowledge points. We provide their corresponding English translations.}
    \label{fig:case}
\end{figure*}
K12 (Grades 1–12) science knowledge is the center of various 21st-century skills  \cite{kennedy202021st}, requiring domain-specific expertise, rigorous logical thinking, and the capacity for multi-step reasoning. It serves as a foundation for solving a wide set of real-world problems, such as coding to solve real-world problems, analysing statistical data, and computing the expenses for a business plan. 
Moreover, there are diverse question types in K12 education, which can more comprehensively and accurately evaluate human knowledge understading and thinking reasoning.
For example, multiple-choice questions focus on information matching, fill-in-blank questions on key information completion, while open-ended questions typically require complex logical reasoning and comprehensive linguistic expression. The combination of these three question types allows for a more thorough evaluation of the model's various capabilities. Consequently, just like humans, the performance boundaries of MLLMs in K12 scenarios stands for their general intelligence capabilities. Systematically exploring the perfermance of MLLMs in k12 Education is crucial for the evaluation of model capabilities.

However, current studies evaluating MLLMs' performance in K12 education surfer from several limitations, including a narrow subject focus, insufficient data scale, and a lack of question types diversity.For instance, MathVista \cite{mathvista}, CMM-Math \cite{liu2024cmm}, and MM-PhyQA \cite{Mm-phyqa} focus on single subjects, while multidisciplinary evaluations like STEM \cite{stem} primarily target elementary levels. Furthermore, existing K12 benchmarks such as GaokaoMM \cite{gaokao} and CMMU \cite{he2024cmmu} hampered by small dataset sizes and a prevalence of multiple-choice questions, with CMMU having 80\% of its questions in this format, thus hindering a comprehensive exploration of MLLMs' capabilities within K12 contexts.
Furthermore, current evaluation methods primarily focus on the accuracy of the final answer, neglecting the assessment of the model's underlying reasoning process. This oversight is particularly pertinent for models like Deekseek-R1 \cite{guo2025deepseek}, which have recently emphasized improving final answers through enhanced Chain-of-Thought (CoT) reasoning. Consequently, a thorough evaluation of their reasoning process is crucial for the advancement of reasoning models.
However, effective methods for evaluating the entire reasoning process of models, as well as metrics for assessing the quality of this process, remain largely unexplored.

%
To address these challenges, we introduce \textbf{K12Vista}, a Chinese scientific subjects benchmark across K12 with three question types. K12Vista comprises 33K questions spanning five core scientific subjects: mathematics, physics, chemistry, biology, and geography. For each question, we provide fine-grained metadata, including grade, question type, knowledge points, difficulty level, and detailed reference solution steps annotations. By enabling categorization across subjects, grades, and question types, K12Vista supports granular analyses of model performance. Moreover, comparing to traditional binary correct/incorrect judgments of model fianl answer, we introduce a novel evaluation method named \textbf{step-by-step evaluation} that employs a process evaluation model namely \textbf{K12-PEM} tailored to our benchmark, to first extract key reasoning steps from the chain-of-thought (CoT) reasoning response of MLLM, then judge the correctness of intermediate steps and answers, classify and analyze errors, and generate an overall score for the entire response. This approach systematically reveals the CoT reasoning quality of MLLMs, overcoming the limitations of superficial evaluation relying solely on final answers. Some examples are shown in Figure \ref{fig:case}.
Meanwhile, we leverage an automated data pipeline to construct \textbf{K12-PEM-800K} a large scale process evaluation multimodal dataset offering detailed step-by-step evaluation annotations for multimodal reasoning process. Fine-tuning on the K12-PEM-800K dataset can significantly enhance the model's ability to evaluate reasoning processes.
We also introduce \textbf{K12-PEBench}, a high-quality, human-annotated benchmark designed to assess the effectiveness of process evaluation.

We evaluated a range of advanced MLLMs on K12Vista. Experimental results demonstrate that Models equipped with reasoning-enhanced capabilities such as Gemini-2-thinking, O3-mini typically demonstrate superior performance. Related analysis reveals notable deficiencies in the multimodal reasoning processes of current MLLMs, providing critical insights for the development of next-generation models.

Our contributions are summarized as follows:
\begin{enumerate}[leftmargin=*]
    \item We present a novel all-encompassing Chinese multimodal benchmark for efficiently evaluate K12 subjects knowledge understanding and reasoning performance across different educational levels and question types.
    \item To enhance models' ability to evaluate CoT reasoning processes, we constructed a massive multimodal process evaluation dataset, K12-PEM-800K. Building upon this dataset, we then introduced K12-PEM, a process evaluation model designed to implement a novel process evaluation method.
    \item We develop K12-PEBench, a high-quality, human-annotated benchmark designed to evaluate the abilities of MLLM-based process evaluation.
    \item We conducted massive experiments and performed a deep analysis of their performance on \ours, providing clear pathways for model optimization.
\end{enumerate}

\begin{table}[!t]
  \centering
  \small
    \caption{\textbf{Comparison between K12Vista and existing K12 related multimodal benchmarks.} K12Vista offers more comprehensive data and question coverage, alongside rich metadata and a step-by-step evaluation method that enable reliable assessments of MLLMs’ CoT reasoning process. Lang:language; KnowPoints: knowledge points; RefSolu: reference soluation; MC: Multiple Choice; FR: free-response; Fill: fill-in-blank.}
  \begin{tabularx}{\textwidth}{@{}
  >{\raggedright\arraybackslash}p{2.7cm} 
  >{\centering\arraybackslash}p{0.6cm} 
  >{\centering\arraybackslash}p{0.6cm} 
  >{\centering\arraybackslash}p{1.5cm} 
  >{\centering\arraybackslash}p{1cm} 
  >{\centering\arraybackslash}p{2cm} 
  >{\centering\arraybackslash}p{1cm}  
  >{\centering\arraybackslash}p{1.5cm}  
  @{}}
\toprule
\multicolumn{1}{l}{\multirow{2}[4]{*}{\normalsize Benchmarks}}  & 
\multicolumn{5}{c}{\normalsize Data Features} & 
\multicolumn{2}{c}{\raisebox{-1ex}{\includegraphics[width=0.5cm]{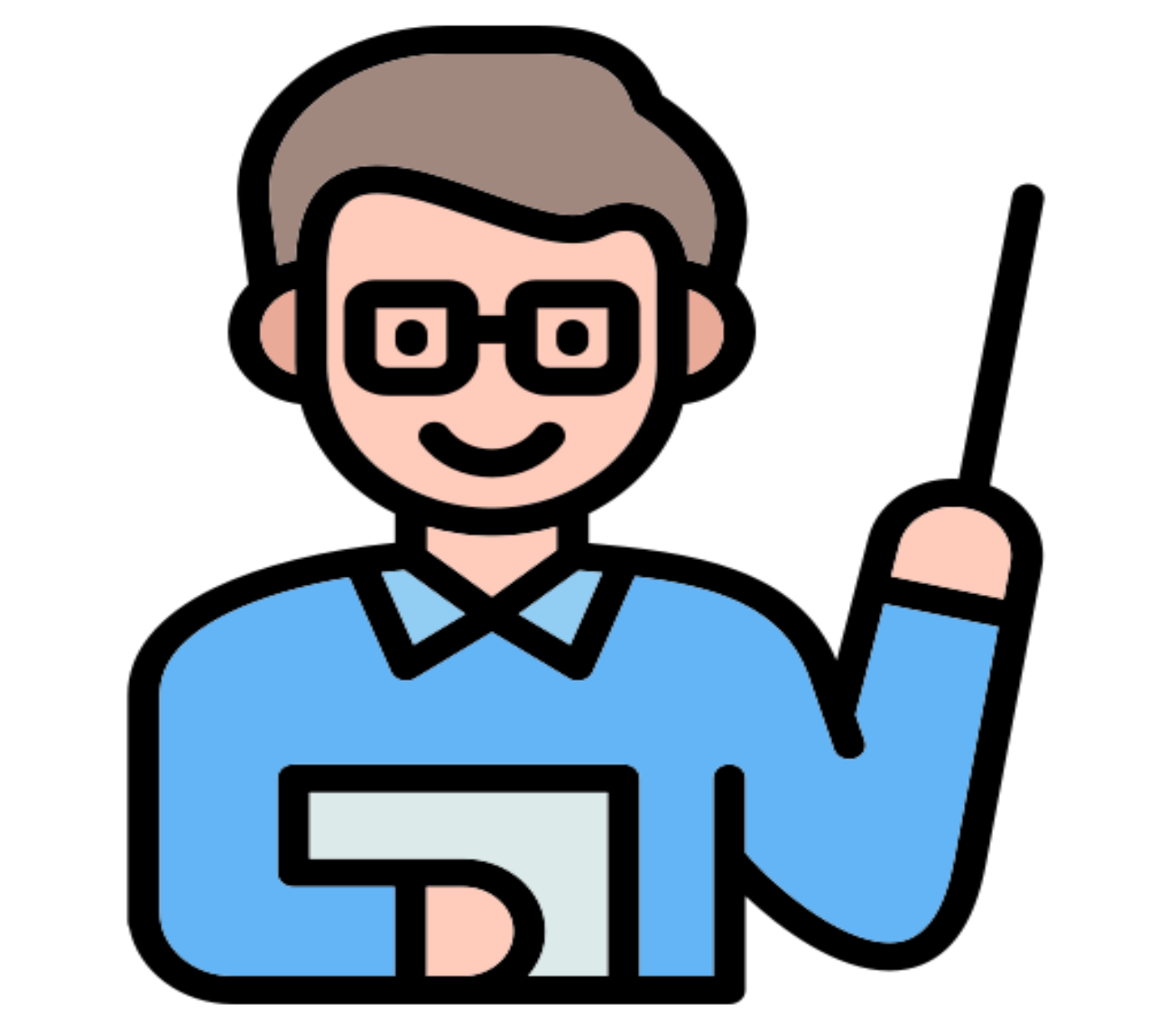}\normalsize} Evaluation}\\
  \cmidrule(l){2-6} \cmidrule(l){7-8}
\multicolumn{1}{c}{} & 
Lang&Num&KnowPoints&Grades&Question Types&RefSolu&ProcessEval\\
    \midrule 
    MM-PhyQA\cite{Mm-phyqa}&en&4.5K&41&K10\(\sim\)12&MC&\xmark&\xmark\\
    CMM-math\cite{liu2024cmm}&zh&28K&13&K1\(\sim\)12&MC,Fill&\cmark&\xmark\\
     Visscience\cite{jiang2024visscience}&en,zh&3.0K&-&K1\(\sim\)12&MC&\xmark&\xmark\\
    STEM\cite{stem}&en&214K&448&K1\(\sim\)K8&MC&\xmark&\xmark\\
    CMMU\cite{he2024cmmu}&zh&3.6K&-&K7\(\sim\)12&MC,Fill&\cmark&\xmark\\
    GAOKAO-MM\cite{gaokao}&zh&0.6K&-&K10\(\sim\)12&MC,FR&\cmark&\xmark\\
    \midrule 
    K12Vista&zh&33K&\textbf{17721}&\textbf{K1\(\sim\)12}&\textbf{MC,Fill,FR}& \cmark &\includegraphics[width=0.85cm]{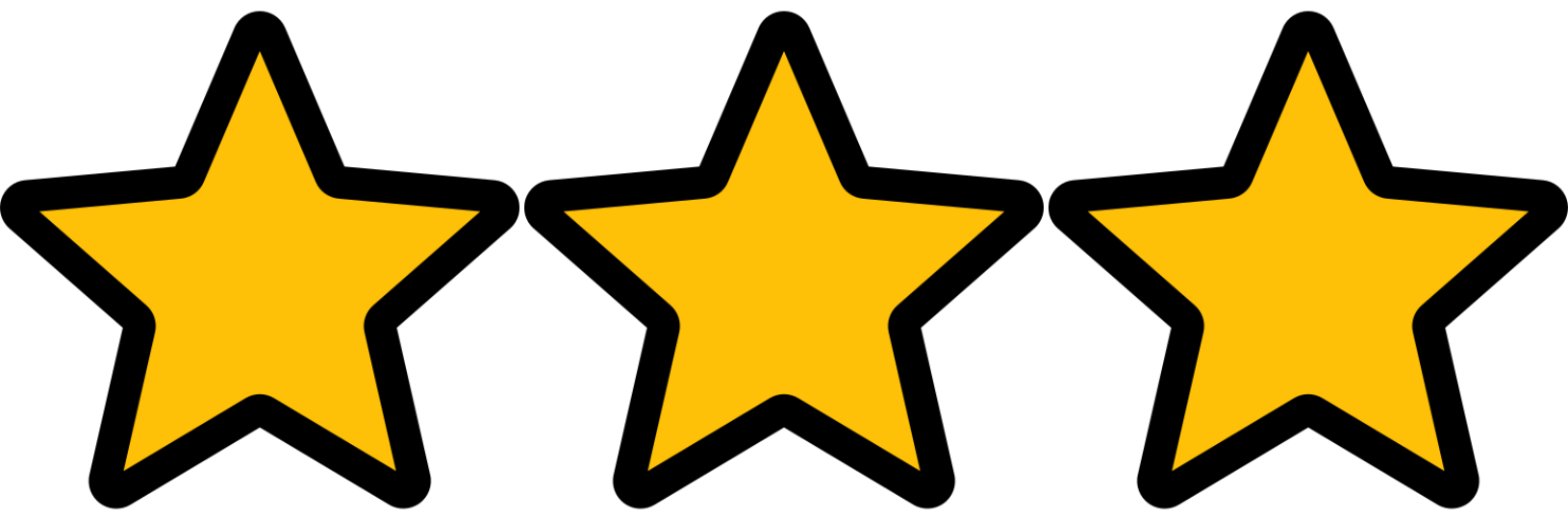}\\
    \bottomrule 
\end{tabularx}
\label{tab:Comparation} 
\end{table}
\section{Related Work}
\subsection{MLLM Benchmark}
With the rapid evolution of MLLMs, a variety of benchmarks have been proposed to assess their performance \cite{okvqa,textvqa,realworldqa,scienceqa,SEEDbench,ocrbench}. However, most benchmarks predominantly focus on basic perceptual skills, falling short of evaluating in depth domain knowledge reasoning or only conducting reasoning in limited contexts. MathVista \cite{mathvista} emphasizes visual-mathematical comprehension, CMM-math \cite{mathverse} focuses on K12 mathematics, MM-PhyQA \cite{Mm-phyqa} centers on high-school physics, while MME and MMBench examine basic visual understanding and cross modality fusion \cite{mme,mmbench}. Recently, more comprehensive evaluations have emerged. For example, MMMU \cite{mmmu} presents a university-level challenge across various academic disciplines. Despite these advancements, existing benchmarks still face limitations in data scale, annotation richness, and question type diversity.
\subsection{MLLM-based Process Judgement}
Process evaluators based on MLLMs have been widely utilized to automatically assess the multimodal reasoning steps of MLLMs \cite{wang2025visualprm,luo2024improve}. Visual-PRM \cite{wang2025visualprm} specifically focuses on the Best-of-N evaluation, enhancing multimodal reasoning performance by scaling the test-time of MLLMs. Additionally, there are already some reasoning benchmarks that employ MLLMs to evaluate the intermediate processes of model outputs. MathVerse leverages GPT-4V \cite{gpt4v} to extract and assess key reasoning steps, providing detailed error analysis and an overall score. OlympicArena \cite{huang2024olympicarena} uses GPT-4V to rate the correctness of each solution step, ensuring a rigorous evaluation. Currently, evaluations mainly rely on closed-source models, which are excessively costly and have unstable reproducibility.
\section{K12Vista}
\subsection{Overview of K12Vista}
\begin{figure*}[t]
    \centering
    \includegraphics[width=1\textwidth]{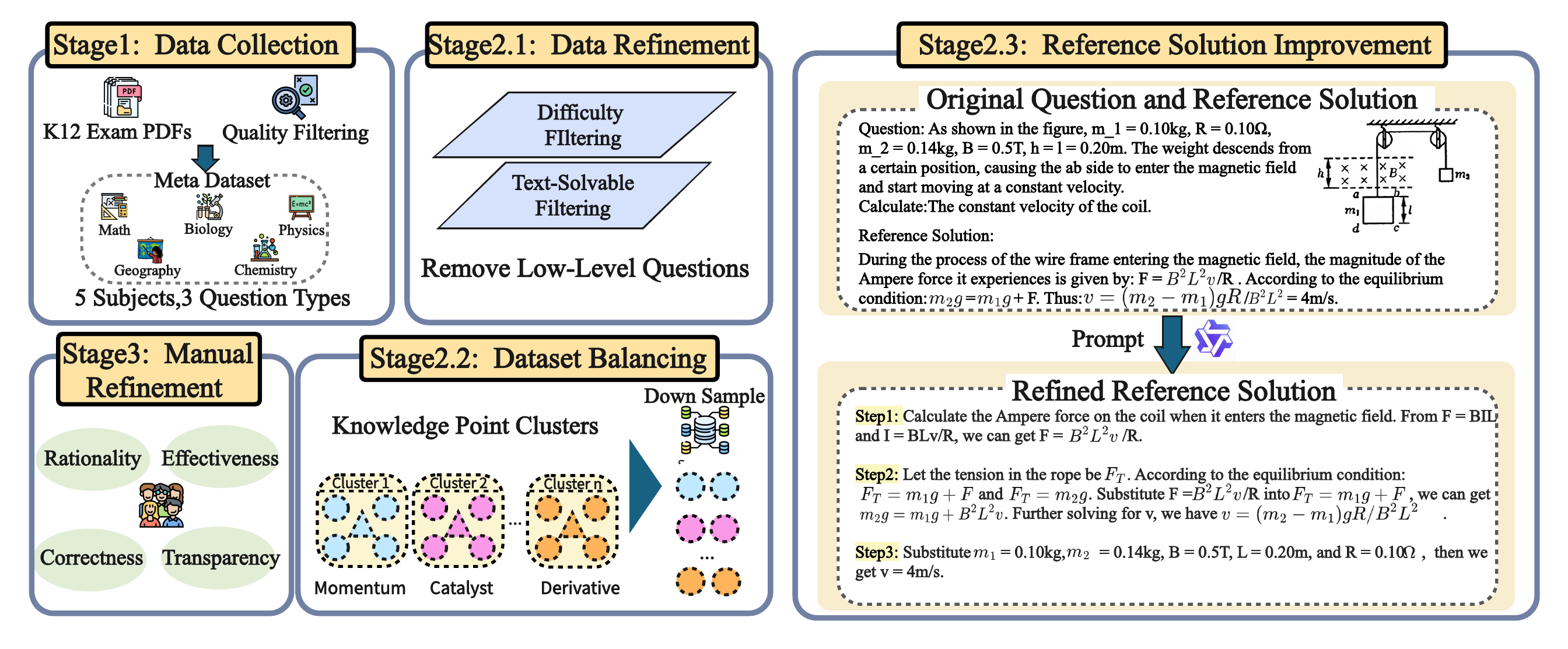}
    \caption{Overview of K12Vista dataset construcion process.}
    \label{fig:data_pipeline}
\end{figure*}

The dataset construction of our benchmark involves three stages, as shown in Figure \ref{fig:data_pipeline}. To mitigate data contamination risks, our dataset primarily sources questions from various non-public offline school exams, rather than textbooks or online question banks.We developed K12Vista by continuously sourcing materials from the non-public school examinations with authorization from the data providers over a six month period. 
K12Vista is a novel Chinese multimodal benchmark designed to assess the comprehension and reasoning capabilities of MLLMs across five scientific disciplines: mathematics, physics, chemistry, biology, and geography. Spanning primary to high school levels, it supports systematic assessment of models' knowledge mastery and reasoning abilities across different educational stages.Our K12Vista contains 3 types of questions: 
1) Multiple Choice Questions: each question provides 4 options with only one correct answer.
2) Fill-in-blank: The model fills in the blanks with the correct answer to complete the sentence or article.
3) Free-response questions: The model uses its knowledge, understanding and thinking skills to respond in writing to the questions posed. As shown in Table \ref{tab:Comparation}, K12Vista offers more comprehensive data and question coverage.

\subsection{Data Construction}
\paragraph{Data Collection} \begin{wraptable}[15]{r}{0.49\textwidth} 
    \centering
    \caption{The statistic of K12Vista} 
    \begin{tabular}{lccccccccccc} 
        \toprule 
        Statistic & \#Number \\
        \midrule 
        Question Number & 33,660 \\
        Total Subjects & 5 \\
        Total Knowledge Points & 17721  \\
        Total Question Types & 3 \\
        \midrule 
        Avg. Problem Tokens & 150.44 \\
        Avg. Reference Solution Tokens & 247.48 \\
        Avg. Reference Solution Steps & 5.15 \\
        Avg. Number of Answers & 3.24 \\
        Avg. Answer Tokens & 24.88 \\
        \bottomrule 
    \end{tabular}
    \label{tab:Statisitc table} 
\end{wraptable} Questions were extracted from original PDF documents, then automatically processed into LaTeX files using the OCR tool Mathpix to retrieve text, which was subsequently converted into JSONL format. Corresponding images were resized to standardized dimensions, while all mathematical and scientific formulas were preserved as native LaTeX notation to maintain structural accuracy. This effort produced a large-scale question bank comprising approximately 300,000 questions, covering the entire K12 educational spectrum, multiple disciplines, diverse knowledge points, and question formats, serving as the metadataset for K12Vista.

Then, we first filtered out blurry images and those with resolutions below the predefined threshold using predetermined rules to ensure image quality standards. Subsequently, we developed a specialized prompt framework based on the Qwen-72B-Instruct model to conduct structural integrity validation on the metadataset: entries with JSON parsing errors, such as missing answer fields, garbled question text, or incomplete metadata were systematically removed. The prompts we use are detailed in Appendix \ref{appendix:Data Construction Detail}. Ultimately, we filtered and obtained approximately 160,000 valid questions.

\paragraph{Data Refinement} To further optimize data quality, we systematically enhanced the dataset by: 1) filtering out low-challenge questions correctly answered by InternVL2-8B, Qwen2-VL-7B, and MiniCPM-V-2.6 to refine difficulty gradients; 2) excluding questions solvable by Qwen2.5-VL-Instruct-72B with text-only inputs to ensure strict multimodal reasoning dependency. 
Subsequently, we clustered questions based on their manually annotated knowledge points, identifying 17,000 core knowledge units. A stratified sampling strategy was subsequently adopted: first, we ensured a minimum sample size of 1,000 questions for each discipline-grade-question type combination, maintaining balanced sample sizes; concurrently, uniform sampling across core knowledge points was implemented, requiring at least one representative instance of each key knowledge point within evaluation subsets (each instance may cover multiple knowledge points) to guarantee comprehensive knowledge coverage. 

\paragraph{Manual validation} \begin{wraptable}[18]{r}{0.49\textwidth} 
\vspace{-1pt}
\setlength{\tabcolsep}{2pt}
    \centering
    \caption{The Statistic of K12-PEM800K and K12-PEBench} 
    \begin{tabular}{lc} 
        \toprule 
        \textbf{Statistic} & \textbf{\#Number} \\
        \midrule 
        \textbf{K12-PEM-800K} & 840,175 \\
        Av g Reference Solution Tokens & 257.40 \\
        Avg. Problem Tokens & 150.59 \\
        Avg. Student Input Tokens & 423.79 \\
        Avg. Output Tokens& 331.04 \\
        Avg. Student Error Steps & 3.99 \\
        \midrule 
        \textbf{K12-PEBench} & 3,033 \\
        Avg. Reference Solution Tokens & 220.34 \\
        Avg. Problem Tokens & 134.43 \\
        Avg. Student Input Tokens& 317.59 \\
        Avg. Student Input Steps & 6.32 \\
        Avg. Student Error Steps & 2.95 \\
        \bottomrule 
    \end{tabular}
    \label{tab:pem_statistic} 
    \vspace{-5pt}
\end{wraptable}

 To ensure the quality of the K12-Vista benchmark, we implemented a rigorous manual verification process across the entire dataset. First, we leveraged Qwen2.5-VL-72B to reconstruct the raw unstructured reference solutions, decomposing them into logically clear structured reasoning steps to form a high-quality, uniformly standardized step-by-step solutions. Then, a validation team of ten senior undergraduate students meticulously reviewed each data item, conducting multidimensional verification of the question content, image, and reconstructed reference solutions to rectify logical fallacies or scientific inaccuracies. This ensured the scientific validity of the reasoning process and the standardization of the solution format, providing high-quality benchmark data for process evaluation evaluation.

\subsection{Data distribution and statistics}
As shown in Table \ref{tab:Statisitc table}, K12Vista comprises five core subjects, including mathematics, physics, chemistry, biology, and geography, each subject featuring three question types: multiple-choice, fill-in-blank, and free-response. With 1,000 questions per type in each subject, the benchmark ensures comprehensive coverage across subjects and question formats to enable rich, multifaceted evaluation of MLLMs' capabilities.

\subsection{Quality Evaluation}
To investigate K12Vista’s quality, we randomly selected 1000 samples for assessment. Three professional data inspectors evaluated them, resulting in high-quality rates of 99\% for questions, 96\% for answers, and 94\% for reference solutions (see Appendix \ref{appendix:appendix_K12Vista_Quality Assessment}). 

\section{Process Evaluation Method}
\subsection{K12-PEM-800K Construction}
\begin{figure}[t]
    \centering
    \includegraphics[width=1\textwidth]{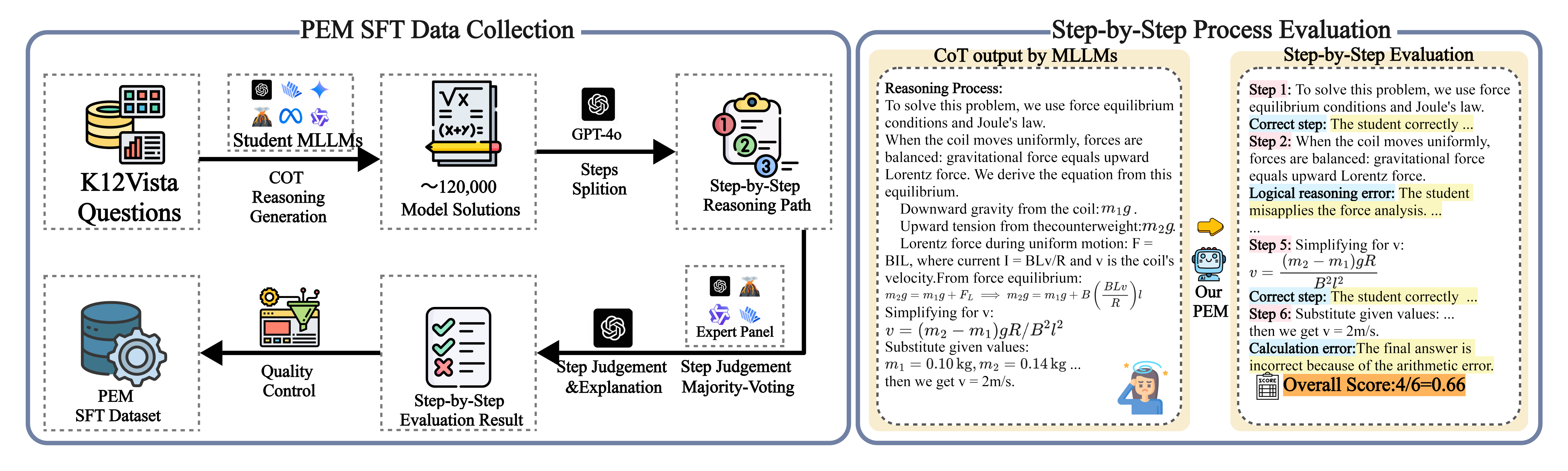}
    \caption{Overview of K12-PEM-800K data construcion process.}
    \label{fig:prm_data_collect}
\end{figure}

\paragraph{MLLM Error Analysis} As shown in Figure \ref{fig:prm_data_collect}, to enable reliable and step-wise evaluation of CoT reasoning processes, we first systematically analyze common error types. We collected CoT solutions for each question in K12Vista from various MLLMs. Based on a comprehensive analysis of errors MLLMs typically make during CoT reasoning, we inductively defined nine step-wise error categories: Image cognition error, Question misunderstanding, Lack of relevant knowledge, Knowledge application error, Logical reasoning error, Hallucination error, Calculation error, and Incomplete answer error. Definitions for each category are provided in Appendix \ref{appendix:Definitions for each category}.
\paragraph{MLLMs' CoT Collection} Then, to replicate the complexities inherent in real-world evaluation scenarios, we leveraged 40 MLLMs of diverse scales including GPT-4o \cite{gpt4}, Internal2.5VL series \cite{Internal25vl}, QwenVL series \cite{qwen25}, and LLaVA-Onevision series \cite{li2024llava},to generate CoT reasoning outputs on K12Vista benchmark. The complete model list and generation prompts are detailed in Appendix \ref{appendix:Models and Prompt for Solution Generation}. 
\paragraph{MLLM Based Step-Wise annotation} These outputs were first decomposed into structured step-wise reasoning paths using GPT-4o, which were then submitted to an expert model panel comprising GPT-4o, Gemini2-Thinking, Qwen2.5-V1-72B, and InternVL2.5-78B-MPO for granular step-level evaluation. The panel operated through a systematic workflow: individual models independently judged the correctness of each reasoning step and labeled error types, with final determinations made via a majority-voting mechanism; GPT-4o subsequently generated step-specific explanations, producing standardized triple-tag list $[s_i, j_i, r_i]$for each step,where is the split reasoning step; j denotes the judgment type;r is the explanations of judgement. The aggregated triple-tag list for each reasoning path constituted its fianl evaluation result list. Please refer to Appendix \ref{appendix:Prompt for K12-PEM-800K Generation} for more details.
\paragraph{Data Filtering} To guarantee data integrity, a dual-filtering protocol was implemented: (1) Format Integrity: Samples deviating from the predefined format (each step must include reasoning step, judgment type, explanation as a list) 
 (2) Explanation Rationality: Samples with unreasonable explanation (e.g., judgment type and explanation are inconsistent). Through this automated data refinement pipeline, we ultimately generate almost 900,000 diverse CoT evaluation samples. We selected 840,175 of these as the final K12-PEM-800K.\begin{wrapfigure}[18]{r}{0.45\textwidth} 
    \centering
    \includegraphics[width=\linewidth]{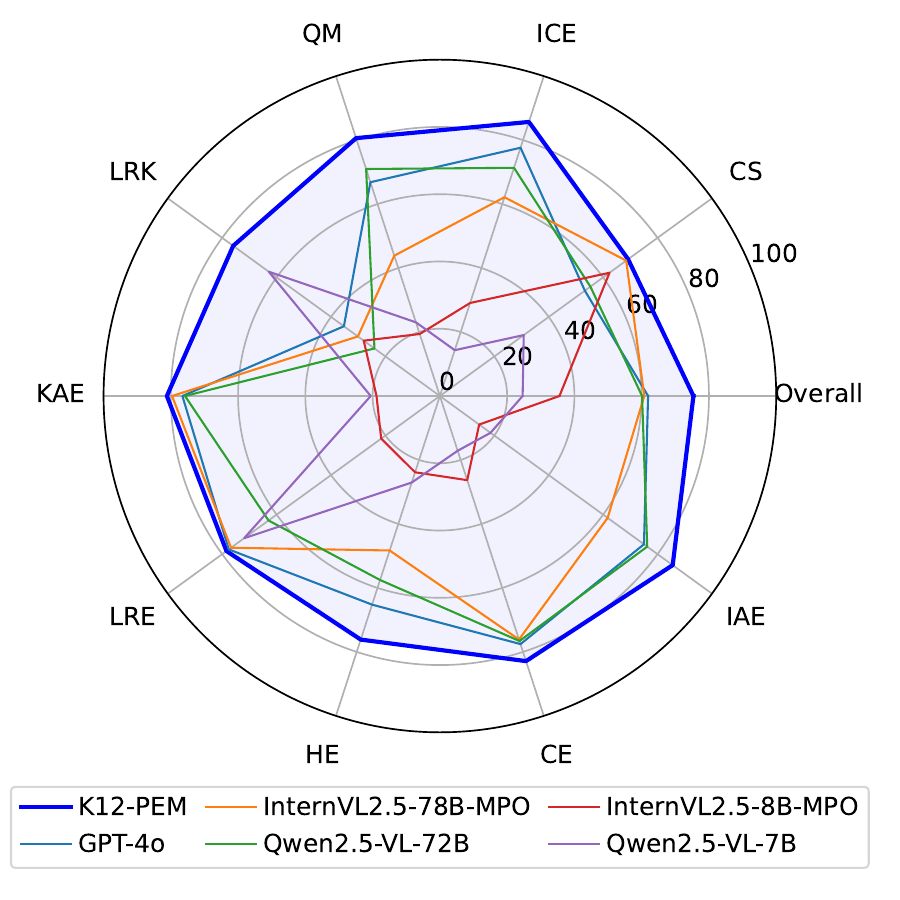} 
    \caption{The result of K12-PEM on K12-PEBench.}
    \label{fig:prm_rad_figure}
\end{wrapfigure} 
\subsection{K12-PEBench Construction}
To construct K12-PEBench, we carefully selected from our previously collected CoT evaluation samples. After excluding the K12-PEM-800K dataset, we chose approximately 3,000 samples with rich reasoning content to form its foundation, with
detailed statistics presented in Table \ref{tab:pem_statistic}. Subsequently, a validation team of ten undergraduate students, who had passed proficiency exams and completed annotation tutorials, performed a second round of manual annotation on these data. The annotators' primary task was to judge the correctness, identify error types, and analyze the root causes for each reasoning step in the CoT solutions. We define the evaluation metrics as the classification accuracy for nine step-level annotation types (comprising eight defined error types and a correct step annotation) and the overall accuracy.

\textbf{Quality Evaluation} To investigate K12-PEBench’s quality, we randomly selected 100 samples for assessment. Three professional data inspectors evaluated them, resulting in high-quality rates of 99\% for questions and student solutions, 90\% for step-wise label (see Appendix \ref{appendix:appendix_K12-PEMBench_Quality Assessment}). 

\subsection{Process Evaluation Model K12-PEM}
We fine-tuned Qwen2.5-VL-7B-Instruct model using the collected K12-PEM-800K train dataset, improving its reasoning quality judgments. We compared it against other candidate models on the K12-PEBench. As shown in Figure \ref{fig:prm_rad_figure}, the result demonstrates that our PEM can accurately reflect the correctness of reasoning steps. Please refer to Appendix \ref{appendix:K12-PEM Detail} and \ref{appendix:K12-PEMBench Result} for detailed information on the SFT phase and specific results on K12-PEBench.

\section{Evaluation}
To establish a comprehensive evaluation framework, we developed two evaluation modalities: direct inference evaluation and CoT reasoning step-by-step evaluation.
\subsection{Direct inference evaluation}
First, we instruct the model to output answers directly without intermediate reasoning steps. Under this modality, the model produces only the final answer without derivational steps. We then use Qwen2.5-VL-72B-Instruct to extract the final answer from the model’s output, compare it against the reference answer, determine correctness, and calculate the final score. The specific content of the infer prompts and answer extraction prompts is detailed in Appendix \ref{appendix:Prompt for MLLMs Infer}. For questions containing multiple sub-questions or answer elements, we count the number of correctly answered components. For example, if a fill-in-the-blank question has two blanks and the model correctly completes one, the score is 0.5. Please refer to Appendix \ref{appendix:Prompt for Result Evaluation} for more details.
\begin{table}[!t]
\caption{Performance of MLLMs across Primary, Middle, and High School Grades under Direct Inference and CoT Reasoning Step-by-Step Evaluation. Overall performance represents the average accuracy across all questions.}
\centering
\resizebox{\textwidth}{!}{%
\begin{tabular}{lcccc|cccc} 
\toprule
\multicolumn{1}{c}{\multirow{2}{*}{\textbf{Model}}} 
& \multicolumn{4}{c}{\textbf{Direct Inference Score}} 
& \multicolumn{4}{c}{\textbf{Step-by-Step Score}} \\
\cmidrule(lr){2-5} \cmidrule(lr){6-9}
& Primary & Middle & High & Overall & Primary & Middle & High & Overall \\
\midrule
\multicolumn{9}{c}{MLLMs: Text + Image as Input} \\
\midrule
Gemini2-thinking&\textbf{60.79}&\underline{57.75}&\textbf{52.02}&\textbf{55.47}&\textbf{62.06}&\textbf{59.52}&\textbf{54.18}&\textbf{57.36}\\
Qwen2.5-VL-32B&\underline{59.03}&\textbf{58.49}&\underline{51.51}& \underline{55.42}&\underline{57.91}&\underline{56.50}&49.17&\underline{53.35}\\
Qwen2.5-VL-72B&54.08&54.72&47.39&51.39&55.27&53.83&44.79&49.93\\
Gemini2-flash&56.70&55.17&45.69&51.08&54.63&51.12&41.95&47.34\\
QVQ-72B-preview&54.99&51.99&45.9&49.54&47.42&48.15&44.17&46.31\\
InternVL2.5-MPO-78B&50.32&48.22&41.55&45.43&50.23&46.15&37.86&42.82\\
InternVL2.5-78B&45.49&43.62&36.04&40.41&47.29&42.10&33.05&38.53\\
GPT-4o&45.56&37.42&30.39&35.02&48.28&37.44&29.80&35.00\\
Qwen2-VL-72B&40.17&37.92&29.74&34.48&32.88&29.32&20.48&25.71\\
LLaVA-OneVision-72B&33.68&34.59&28.01&31.57&32.70&30.10&22.88&27.11\\
Qwen2.5-VL-7B&40.40&44.97&34.32&39.82&38.67&30.90&20.89&27.16\\
InternVL2.5-MPO-8B&33.12&33.41&25.68&29.94&35.37&30.46&21.53&26.93\\
InternVL2.5-8B&28.90&30.53&23.63&27.31&29.89&25.88&17.93&22.69\\
Qwen2-VL-7B&31.22&28.47&21.71&25.70&18.74&14.62&9.21&12.58\\
\midrule
\multicolumn{9}{c}{LLMs: Text + Captions as Input} \\
\midrule
O3-mini&56.64&52.25&48.89&51.14&57.49&52.95&\underline{50.05}&52.06\\
Deepseek-v3&52.07&47.82&40.60&44.97&57.62&50.38&42.69&47.60\\
O1-mini&55.02&46.38&39.07&43.89&56.19&45.66&38.75&43.51\\
\bottomrule
\end{tabular}
}
\label{tab:main_table}
\vspace{-1.25pt}
\end{table}
\subsection{CoT reasoning step-by-step evaluation}
In the step-by-step evaluation mode, where the model directly evaluates the student's entire CoT output $x_S$ based on the problem $x_P^j$ and its ground truth $x_A^j$ and reference solution $x_R^j$. For each CoT output, the model decomposes the student output $x_S$ into individual steps, with each step $s_i$ encompassing both the reasoning process and the answer to any sub-problem, organizing them into a list where each element takes the form $[s_i, j_i, r_i]$ representing the step description, judgment, and explanation, respectively. The final structured output is formalized as: $[[s_i, j_i, r_i]_{i = 1}^M]$, $M$ denotes the total number of reasoning steps. Let $N$ be the count of steps in these $M$ steps whose judgment is marked as correct. We define Step-by-Step Score as: $score=N/M$. This score equally weights each reasoning step and sub-problem answer, thereby reflecting the quality of both the reasoning process and the final answer. Please refer to Appendix \ref{appendix:Prompt for Result Evaluation} for more details.
\subsection{Evaluation Quality Assessment}
To investigate evaluation quality, we randomly selected 1000 samples for assessment. Five experts rated Qwen2.5-VL-7B outputs on a 0-1 scale across two evaluation modes. The kappa coefficient between MLLM and expert evaluations were 0.88 for direct inference evaluation and 0.75 for step-by-step evaluation, highlighting the effectiveness of our evaluation method and metrics, even for smaller models. Further details are in Appendix \ref{appendix:evaluation_Quality Assessment}.

\section{Experiments}
\subsection{Baselines and Setting}
We evaluated a range of closed-source and open-source models, including: 1) \textbf{Closed-source models}: GPT-4o \cite{gpt4}, Gemini2-flash \cite{team2023gemini}, Gemini2-thinking \cite{team2023gemini}, O3-mini \cite{Gpt-o3-mini}, O1-mini \cite{Gpt-o1-mini}; and 2) \textbf{Open-source models}: Qwen2.5-VL\cite{yang2024qwen2}, InternVL2.5\cite{Internal25vl}, QVQ-72B-preview \cite{Qvq}, InternVL2.5-MPO \cite{wang2024enhancing}, etc. To evaluate LLMs which only accept text input, we generated captions for image inputs with Qwen2.5-VL-72B model (captioning details are provided in Appendix \ref{appendix:Prompt for caption}) and  concatenated captions with questions as LLM inputs. Closed-source models were evaluated via their official APIs, while open-source models were assessed using VLLM on NVIDIA H200 GPUs with default VLLM parameters. See Appendix \ref{appendix:More Result about K12-Vista} for more MLLMs' result. 
\begin{figure}[t]
    \centering
    \includegraphics[width=1\textwidth]{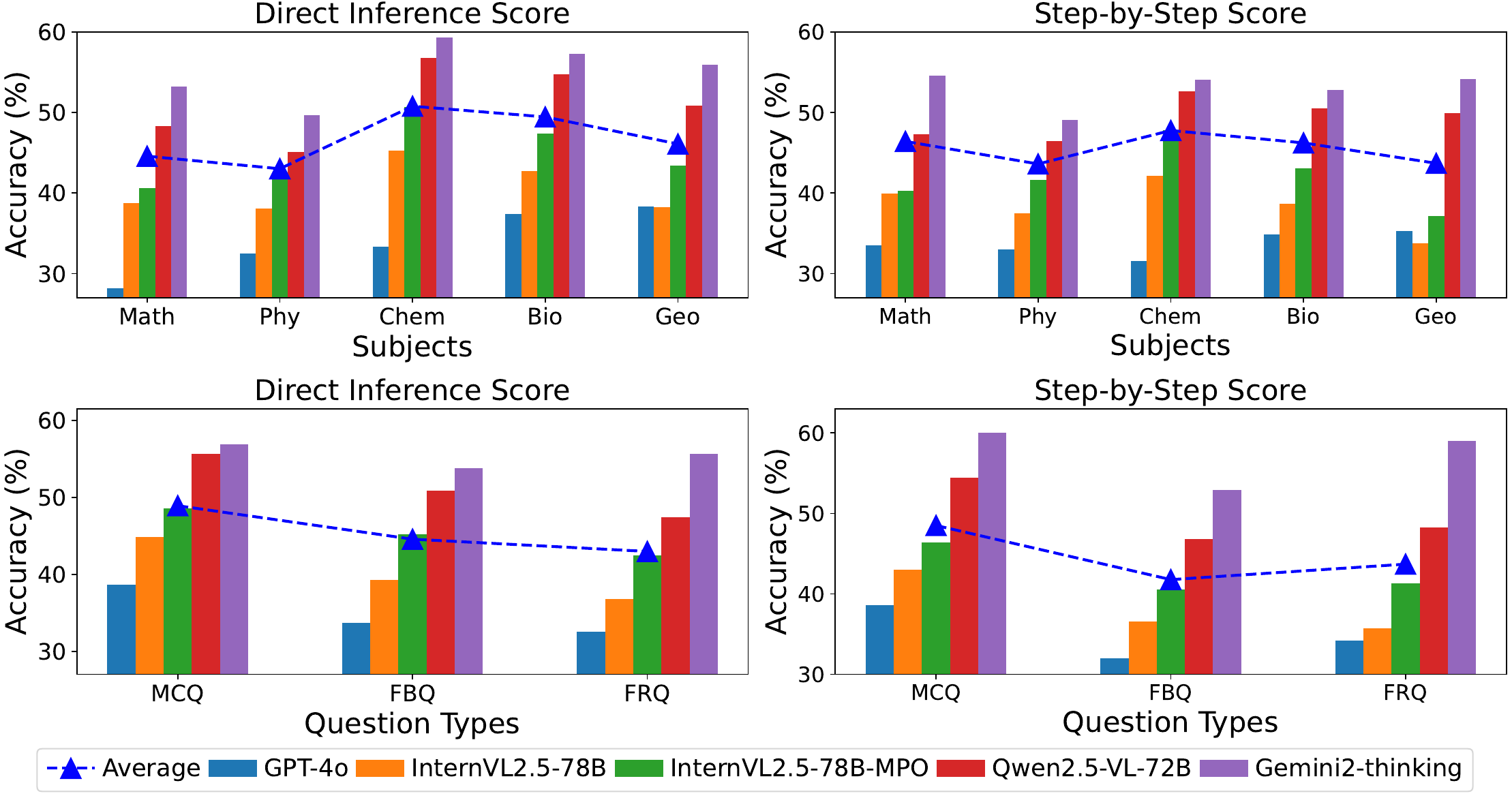}
    \caption{\textbf{Accuracy of MLLMs across different subjects and question types.} We demonstrate the results across five subjects (mathematics, physics, chemistry, biology, and geography) and three question types (multiple-choice, fill-in-blank, free response).}
    \label{fig:figure_questiontype_3}
\end{figure}
\subsection{Main Results}
Table \ref{tab:main_table} presents baseline model performance across primary, middle, and high school grade levels using both Direct Inference and CoT step-by-step evaluation. Gemini2-thinking consistently achieves the highest accuracy in both settings, with overall scores of 55.47\% and 57.36\% respectively, showcasing its superior capability in complex multimodal understanding and reasoning. Qwen2.5-VL-32B follows closely in Direct Inference with 55.42\% but shows a slight drop in Step-by-Step evaluation with 53.35\%, suggesting weaker reasoning process performance. InternVL2.5-8B generally underperforms other models. Overall, larger models tend to perform better.

A consistent trend reveals decreasing accuracy at higher grade levels across all MLLMs, in both evaluation modes, highlighting the increasing demand for deeper understanding and reasoning. The CoT step-by-step evaluation mode particularly challenges MLLMs, effectively differentiating reasoning proficiency. This mode generally yields lower scores for most models, except for reasoning-enhanced ones like Gemini2-thinking and O3-mini, suggesting unenhanced models struggle with reasoning steps. Additionally, LLMs generally perform worse than top MLLMs, underscoring the critical role of visual information in K12Vista.

\paragraph{Results across Different Question Types}
The lower half of Figure \ref{fig:figure_questiontype_3} illustrates accuracy distribution across fill-in-blank (FBQ), multiple-choice (MCQ), and free-response (FRQ) question types. All models score below 60\% in all three types, indicating our benchmark's challenging nature and its effectiveness in identifying model weaknesses. For all models, FBQs consistently yield lower scores than MCQs. This is because FBQs demand complex knowledge integration and generation, whereas MCQs only require selection from predefined answers, making FBQs generally more challenging. Furthermore, the performance gap between models is wider in FRQs than in FBQs and MCQs. This is attributed to FRQs' higher complexity, which necessitates intricate logical reasoning, step-by-step derivation, and comprehensive content generation. Such demands highlight FRQs' superior ability to differentiate model capabilities.
%
%
%
%


 \begin{figure}[t]
    \centering
    \includegraphics[width=1\textwidth]{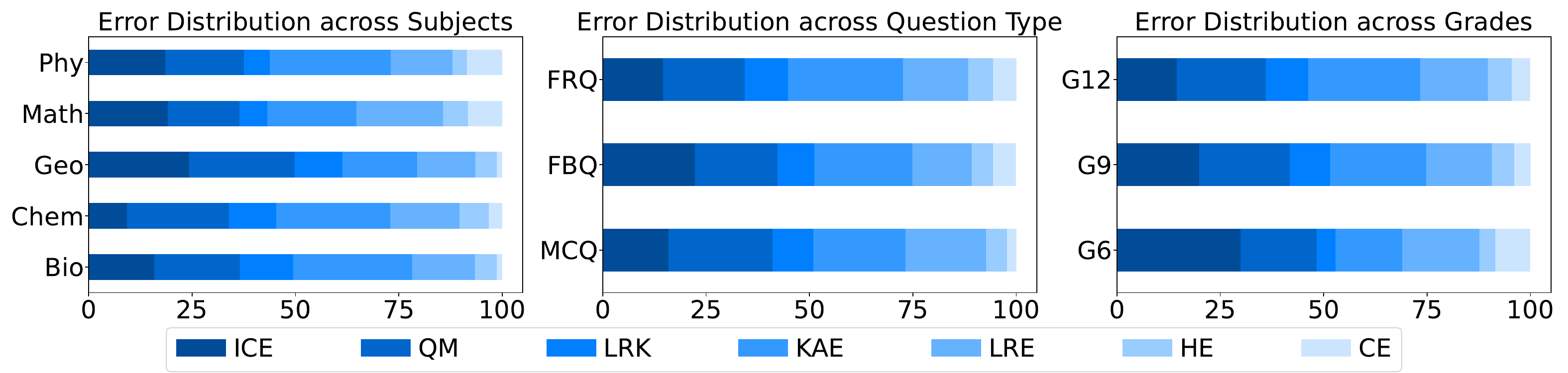}
    \caption{\textbf{Distribution of Step-Wise Error Types} We present the results across various subjects, question types, and grade levels, derived from Gemini2-thinking's result on K12Vista. (ICE: Image Cognition Error; QM: Question Misunderstanding; LRK: Lack of Relevant Knowledge; KAE: Knowledge Application Error; LRE: Logical Reasoning Error; HE: Hallucination Error; CE: Calculation Error;)}
    \label{fig:figure_combined_pro_3}
\end{figure}
\paragraph{Results across Different Subjects}
The upper part of Figure \ref{fig:figure_questiontype_3} shows accuracy distribution across different subjects. Chemistry, Biology, and Geography generally see relatively superior performance from most models under direct inference. For instance, Qwen2.5-VL-72B performs better in these subjects compared to Mathematics and Physics. We attribute this to the former subjects' reliance on factual knowledge and rule-based reasoning, with a greater emphasis on memorization, making them easier for models. In contrast, Mathematics and Physics involve more abstract concepts, logical deduction, and quantitative analysis, demanding complex multi-step reasoning and real-world interpretation, posing a greater challenge. These discrepancies highlight how subject characteristics and knowledge complexity influence model performance, emphasizing the need for subject-specific MLLM benchmarks.
\paragraph{Step-wise error Analysis}
Figure \ref{fig:figure_combined_pro_3} illustrates a distributional analysis of step-wise errors generated by Gemini2-thinking during Step-by-Step Evaluation, revealing significant variations across subjects, question types, and grade levels. At the subject level, Geography shows a notably higher proportion of image perception and text understanding errors due to complex, detail-rich images (e.g., isobaric charts, topographic maps) and substantial background text, which increases comprehension difficulty, while Other subjects primarily concentrate errors in knowledge application and logical reasoning. Mathematics and Physics exhibit slightly more image perception and text understanding errors than Biology and Chemistry, possibly due to prevalent geometry problems (e.g., mechanics diagrams, spatial geometry) and higher computational demands, leading to more calculation errors. At the question type level, Fill-in-Blank Questions show a relatively larger proportion of image perception errors, as they often contain multiple sub-questions and require detailed examination of image information. At the grade level, image and text perception errors significantly decrease with declining grade levels. Conversely, logical reasoning errors, including knowledge deficiency, insufficient knowledge application, and flaws in the reasoning process itself, gradually increase.
\section{Conclusion}
In this paper, we introduce \ours, a novel multidisciplinary Chinese multimodal benchmark, designed to evaluate the understanding and reasoning capabilities of MLLMs on high-difficulty problems in Chinese K-12 core science subjects. Curated from extensive metadata repositories, \ours comprises 33,000 high-complexity questions spanning 12 grade levels, 5 core science subjects, and three question types, enabling comprehensive evaluation and addressing critical gaps in existing benchmarks—including limited data scale, narrow domain coverage, monotonous question formats, and insufficient difficulty.
Additionally, we introduce step-by-step evaluation metrics: by training a special Process Evaluation Model on our benchmark, we enable fine-grained assessment of models’ multi-step reasoning processes, facilitating deeper insights into their performance. Our experiments reveal that current MLLMs face significant challenges in solving complex K-12 problems, particularly exhibiting numerous issues during reasoning. Future work could focus on enhancing the step-wise correctness of models of complex inference, laying the groundwork for more robust multimodal reasoning systems.
\bibliography{main}
\bibliographystyle{plain}

\newpage
\begin{center}
  \Large\bfseries \ours: Exploring the Boundaries of MLLMs in K-12 Education
  \\[1em]
  \ Supplementary Materials
\end{center}
\setcounter{figure}{6}
\setcounter{table}{4}
\appendix
\section{K12Vista Detail}
\subsection{Data Construction Detail}\label{appendix:Data Construction Detail}
\begin{figure}[h]  
    \centering
    \includegraphics[width=1\textwidth]{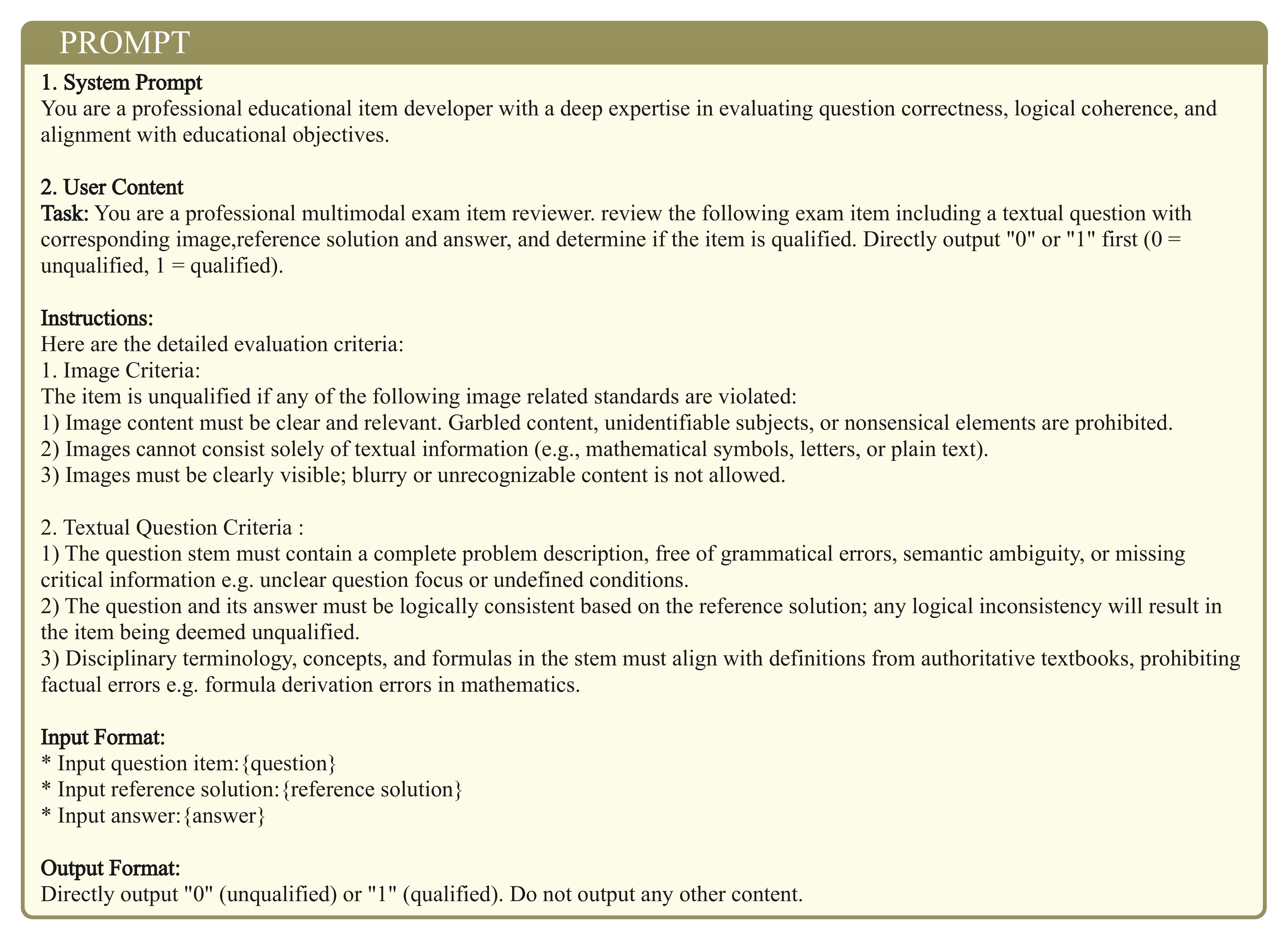}
    \caption{Prompt for question item reviewing. We mainly checked issues with images and question text. We provide their corresponding English translations.}
    \label{fig:appendix_question_item_reviewing}
\end{figure}
To significantly conserve human resources and streamline our workflow, we primarily leveraged Multimodal Large Language Models (MLLMs) for the critical task of data inspection. This automated approach was instrumental in ensuring the quality and integrity of our dataset while minimizing the intensive manual effort typically required.

\begin{wraptable}[13]{r}{0.49\textwidth}
\caption{The High-Quality Rate of 1000 selected Samples} 
\centering
\resizebox{0.5\textwidth}{!}{%
\begin{tabular}{lccc} 
        \toprule 
        \textbf{Set} & \textbf{Question}& \textbf{Answer}& \textbf{RefeSolu} \\
        \midrule 
        \textbf{Math} & 99.50  &99.50 &96.00\\
        \textbf{Physics} & 100.00  &98.00 &96.00\\
        \textbf{Chemistry} & 98.00  &93.50 &92.00\\
        \textbf{Biology} & 99.00  &95.00 &94.00\\
        \textbf{Geography} & 98.50  &94.00 &92.00\\
        \textbf{Overall} & 99.00  &96.00 &94.00\\
        \bottomrule 
    \end{tabular}}
    \label{tab:appendix_k12vista_check} 
\end{wraptable}

Our methodology for this MLLM-driven data inspection involved a systematic, iterative process. We began by thoroughly analyzing a carefully selected sample dataset to identify and categorize prevalent data issues. This initial qualitative assessment allowed us to gain a granular understanding of the types of errors, inconsistencies, or irrelevant content present. Based on these insights, we meticulously designed a series of filtering prompts specifically engineered to detect these identified issues. These prompts were not static; instead, they underwent continuous and iterative refinement. Through repeated testing and adjustment against subsets of the data, we optimized their effectiveness in accurately flagging problematic entries. The culmination of this iterative development, resulting in our most effective and robust prompt, is visually represented and detailed in Figure \ref{fig:appendix_question_item_reviewing}, showcasing its final structure and functionality. This rigorous process ensured that our MLLM-based inspection system was highly precise and efficient in identifying relevant data anomalies.
\subsection{Quality Assessment}\label{appendix:appendix_K12Vista_Quality Assessment}
We randomly sampled 200 questions from each subject, ensuring coverage of different grades. Three data reviewers verified the logical consistency of the questions, the correctness of answers, and whether the reference solutions represented valid problem-solving approaches. The inspection results are presented in Table \ref{tab:appendix_k12vista_check}, demonstrating high-quality rates of 99\% for questions, 96\% for answers, and 94\% for reference solutions.
\section{Evaluation}
\subsection{Prompt for MLLMs Inference}\label{appendix:Prompt for MLLMs Infer}
\begin{figure}[h]  
    \centering
    \includegraphics[width=1\textwidth]{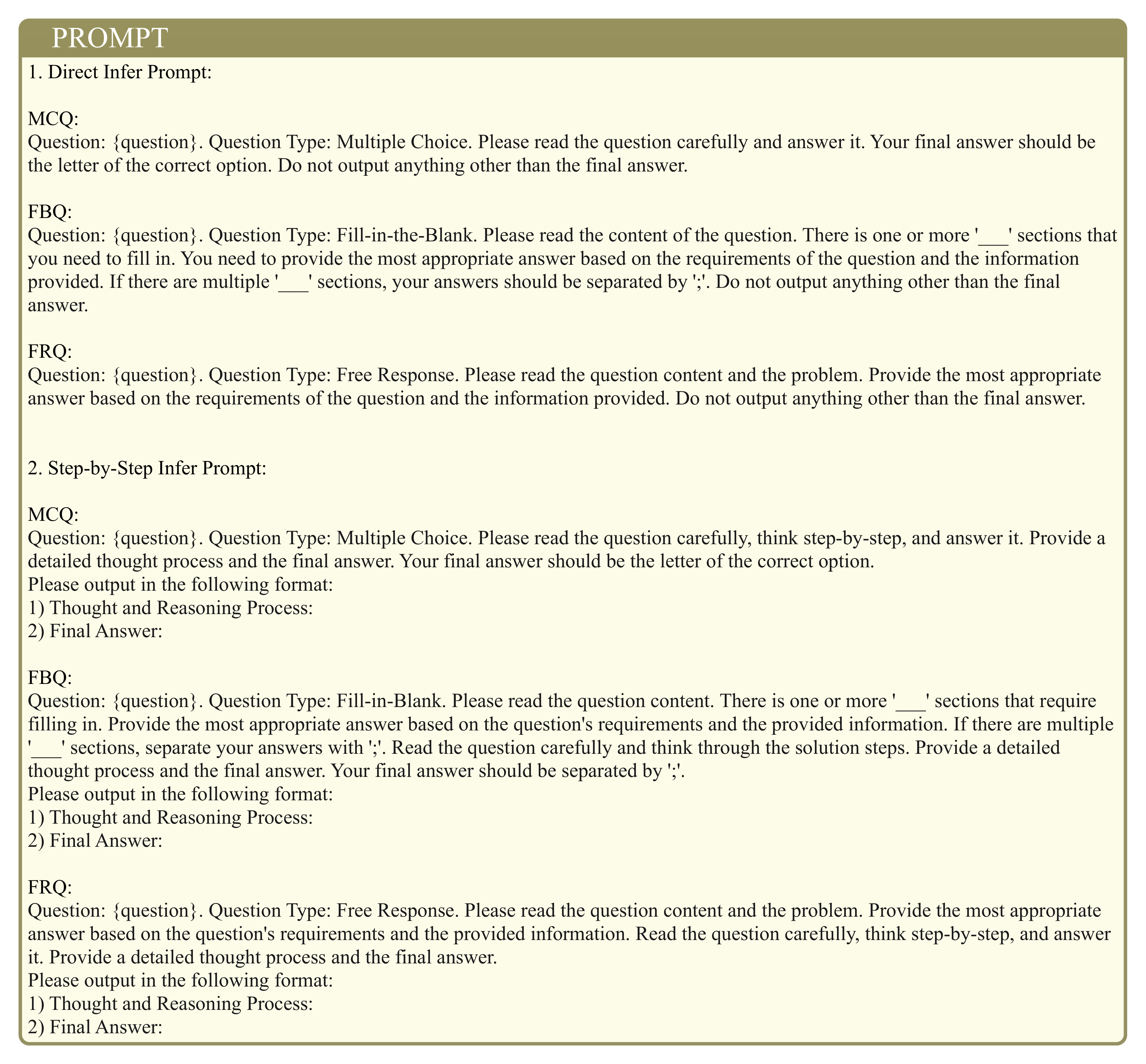}
    \caption{Prompt for MLLMs inference. We designed six distinct prompts for MLLM inference, tailored to different reasoning modes and question types. We provide their corresponding English translations.}
    \label{fig:appendix_inferprompt}
\end{figure}

To comprehensively address the unique challenges posed by various question types across different inference modes, we meticulously designed a dedicated system of prompts. This prompt system is primarily categorized into two core modes: Direct Inference and Step-by-Step Inference. Within each main category, we further refined and customized the prompts, creating specialized versions for three distinct question types: Multiple-Choice Questions, Fill-in-Blank Questions, and Free-Response Questions. This layered and customized design ensures that the model receives the most precise and effective instructions when tackling exams of varying task types and reasoning complexities, thereby maximizing its latent capabilities and guiding it to produce outputs that adhere to the expected format and content.
Figure \ref{fig:appendix_inferprompt} illustrates this in detail.
\subsection{Prompt for Caption}\label{appendix:Prompt for caption}
\begin{figure}[h]  
    \centering
    \includegraphics[width=1\textwidth]{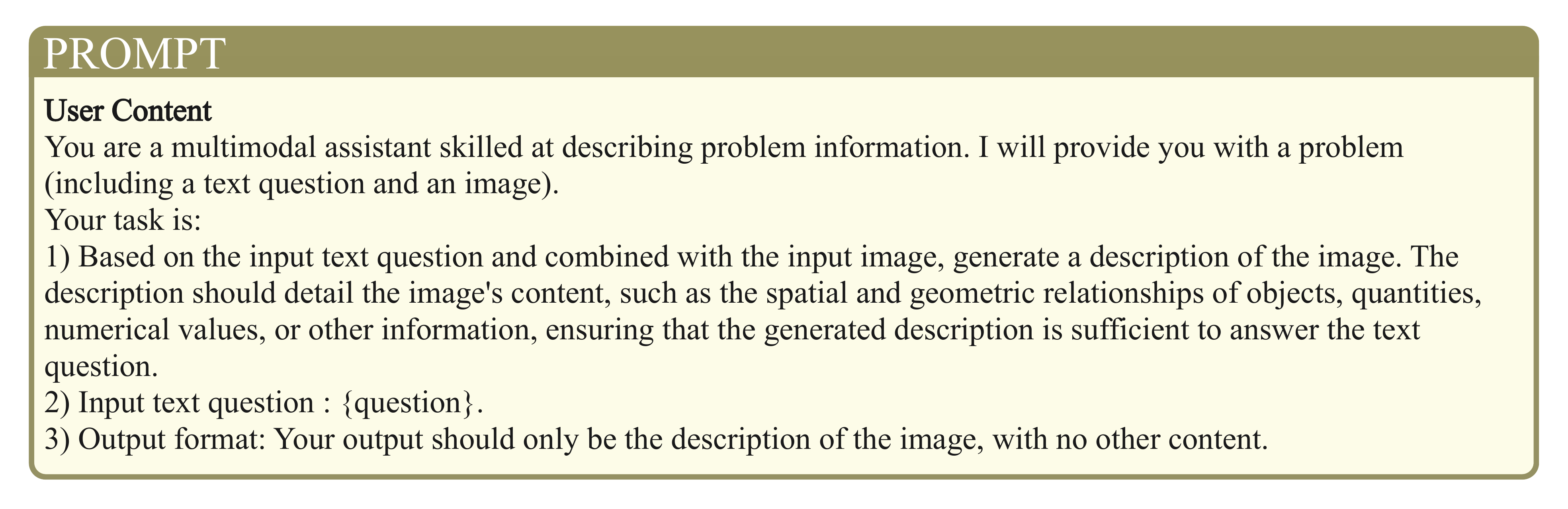}
    \caption{Prompt for image caption. To facilitate LLM inference, we transformed image content into textual representations. We provide their corresponding English translations.}
    \label{fig:appendix_captionprompt}
\end{figure}
To more comprehensively assess the generality and efficacy of our proposed K12Vista framework across different model capabilities, we not only tested Multimodal Large Language Models (MLLMs) but also conducted extended evaluations on purely text-based Large Language Models (LLMs) using the K2vista dataset.
Our specific implementation strategy involved first leveraging the powerful Qwen2.5-VL-72B visual language model to generate detailed textual captions for all image content within the K2vista dataset. Subsequently, we intelligently fused these high-quality image descriptions with the original textual content of the questions. This combined input then served as the processing object for the pure text LLM. This method effectively transformed visual information into a text-comprehensible format, thereby allowing pure text LLMs to indirectly "perceive" and utilize image information for reasoning. The specific prompt used to generate these image captions is detailed in Figure \ref{fig:appendix_captionprompt}, ensuring the reproducibility and transparency of our methodology. Through this approach, we were able to conduct an in-depth analysis of pure text LLMs' performance in this "indirect multimodal" scenario, further validating the comprehensive nature of the K12Vista evaluation framework.
\subsection{Prompt for Result Evaluation}\label{appendix:Prompt for Result Evaluation}
\begin{figure}[!t]  
    \centering
    \includegraphics[width=1\textwidth]{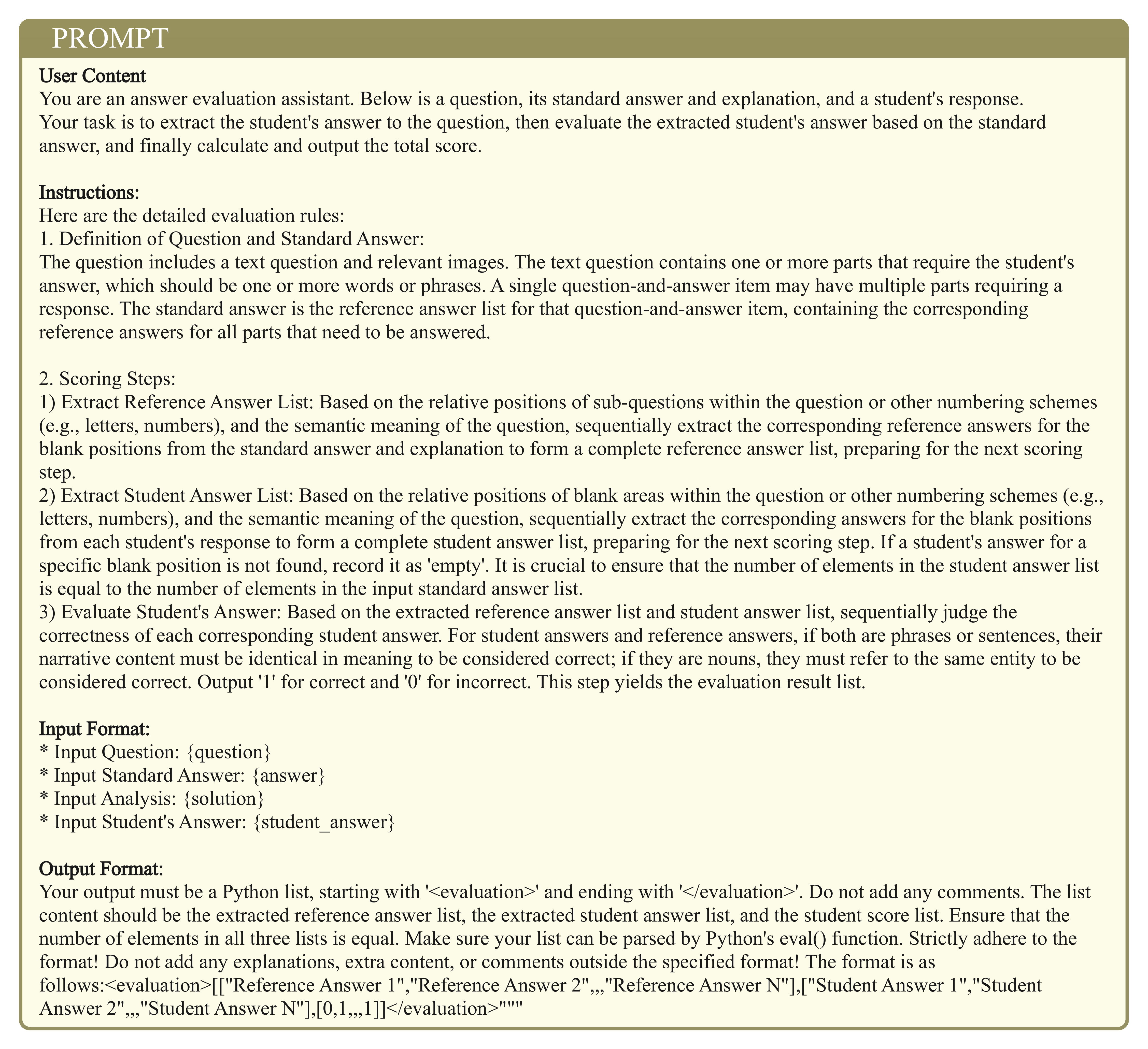}
    \caption{Prompt for direct inference evaluation. We require the MLLM to first extract reference answers, then retrieve students' answers, and subsequently generate a score list through one-to-one comparison. We provide their corresponding English translations.}
    \label{fig:appendix_evalprompt}
\end{figure}

\textbf{Direct inference evaluation}. To thoroughly and precisely assess the capabilities of MLLMs in understanding and answering complex questions, we developed and implemented a precise evaluation strategy. This method aims to meticulously evaluate the model's performance by breaking down the assessment into distinct stages. First, in the Question Comprehension and Intent Identification phase, the model's primary task is to accurately understand the key content required by the question, leveraging the provided original question, standard answer, and detailed explanation. Next, in the Answer Extraction phase, once the model accurately comprehends the question, it proceeds to precisely process the student's free-text responses. This involves accurately extracting the specific answers for each sub-question from the student's response, a process that demands robust information extraction capabilities from the model. Finally, in the Sub-Answer Scoring phase, for each extracted sub-answer, we perform an independent, binary scoring. If the student's sub-answer is consistent in content and semantics with the standard answer, it's judged as correct and assigned 1 point; conversely, any deviation or error results in it being judged as incorrect and assigned 0 points. The average score of these sub-answers is then calculated to provide a comprehensive evaluation. The specific prompt used to direct inference evaluation is detailed in Figure \ref{fig:appendix_evalprompt}

\begin{figure}[!t]  
    \centering
    \includegraphics[width=1\textwidth]{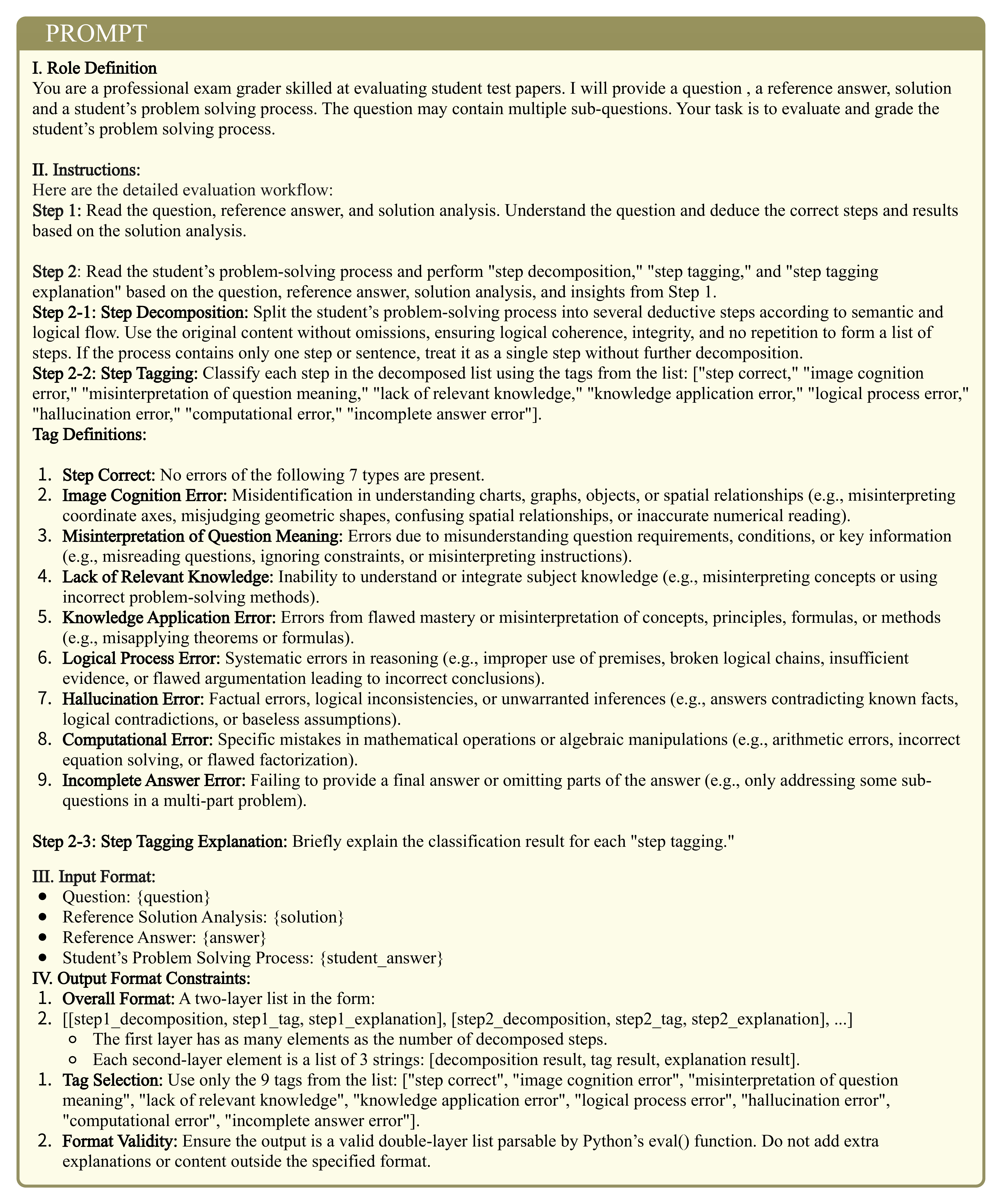}
    \caption{Prompt for step-by-step evaluation. We require K12-PEM to first decompose students' responses into steps and then label each step with one of nine predefined judgement according to the definition. We provide their corresponding English translations.}
    \label{fig:appendix_sbsevalprompt}
\end{figure}

\textbf{Step-by-Step evaluation}. In the step-by-step evaluation mode, where the model directly evaluates the student's entire CoT output $x_S$ based on the problem $x_P^j$ and its ground truth $x_A^j$ and reference solution $x_R^j$. For each CoT output, the model decomposes the student output $x_S$ into individual steps, with each step $s_i$ encompassing both the reasoning process and the answer to any sub-problem, organizing them into a list where each element takes the form $[s_i, j_i, r_i]$ representing the step description, judgment, and explanation, respectively. The final structured output is formalized as: $[[s_i, j_i, r_i]_{i = 1}^M]$, $M$ denotes the total number of reasoning steps. Let $N$ be the count of steps in these $M$ steps whose judgment is marked as correct. We define Step-by-Step Score as: $score=N/M$. This score equally weights each reasoning step and sub-problem answer, thereby reflecting the quality of both the reasoning process and the final answer. Please refer to Figure \ref{fig:appendix_sbsevalprompt} more details.
\subsection{Quality Assessment}\label{appendix:evaluation_Quality Assessment} 
\begin{wraptable}[10]{r}{0.49\textwidth}
\caption{The kappa coefficient between expert evaluations and various evaluation modes} 
\centering
\resizebox{0.5\textwidth}{!}{%
\begin{tabular}{lccc} 
        \toprule 
        \textbf{Evaluation Modes} & \textbf{Direct Score}& \textbf{Step-by-Step Score} \\
        \midrule 
        \textbf{Avg.Expert} & 1  &1\\
        \textbf{Qwen2.5-VL-72B} & 0.88  &0.67\\
        \textbf{GPT-4o} & 0.91  &0.73\\
        \textbf{InternVL2.5-MPO-78B} & 0.83  &0.62\\
        \textbf{K12-PEM} & 0.79  &0.75\\
        \bottomrule 
    \end{tabular}}
    \label{tab:appendix_evaluation_recheck} 
\end{wraptable}

To comprehensively evaluate the performance of models under two evaluation modes, we used GPT-4o, Qwen2.5-VL-72B, and InternVL2.5-MPO-78B for evaluation, respectively. Meanwhile, we invited 5 experts to provide evaluation under the same two evaluation modes and calculated the average. By computing the Kappa coefficient, we found that the two proposed evaluation metrics showed a high degree of consistency with human experts' evaluations, with detailed results listed in Table \ref{tab:appendix_evaluation_recheck}. Finally, considering both cost and accuracy, we decided that Qwen2.5-VL-72B would be used for the Direct Inference evaluation mode, while K12-PEM would be applied to the Step-by-Step evaluation. The Kappa coefficient scoring criteria are as follows: below 0.2 indicates slight agreement, 0.21–0.40 indicates fair agreement, 0.41–0.60 indicates moderate agreement, 0.61–0.80 indicates substantial agreement, and 0.81–1.00 indicates almost perfect agreement.
%
\section{K12-PEM-800K Detail and More Result}
In this section, we provide detailed information on K12-PEM-800K, including the list of MLLMs used for generating solutions
in data construction, data construction process, and more result about experiment.
\subsection{Definitions for each category}\label{appendix:Definitions for each category}
After analyzing numerous instances, we've summarized the following 8 frequently occurring errors and their definitions:

(1) Image Cognition Error: Misidentification in understanding charts, graphs, objects, or spatial relationships (e.g., misinterpreting coordinate axes, misjudging geometric shapes, confusing spatial relationships, or inaccurate numerical reading).

(2) Question Misunderstanding: Errors due to misunderstanding question requirements, conditions, or key information (e.g., misreading questions, ignoring constraints, or misinterpreting instructions).

(3) Lack of Relevant Knowledge: Inability to understand or integrate subject knowledge (e.g., misinterpreting concepts or using incorrect problem-solving methods).

(4) Knowledge Application Error: Errors from flawed mastery or misinterpretation of concepts, principles, formulas, or methods (e.g., misapplying theorems or formulas).

(5) Logical Reasoning Error: Systematic errors in reasoning (e.g., improper use of premises, broken logical chains, insufficient evidence, or flawed argumentation leading to incorrect conclusions).
\begin{table}[!t]
\caption{Performance of MLLMs across Primary, Middle, and High School Grades under Direct Inference and CoT Reasoning Step-by-Step Evaluation. Overall performance represents the average accuracy across all questions.}
\centering
\resizebox{\textwidth}{!}{%
\begin{tabular}{lcccc|cccc} 
\toprule
\multicolumn{1}{c}{\multirow{2}{*}{\textbf{Model}}} 
& \multicolumn{4}{c}{\textbf{Direct Inference Score}} 
& \multicolumn{4}{c}{\textbf{Step-by-Step Score}} \\
\cmidrule(lr){2-5} \cmidrule(lr){6-9}
& Primary & Middle & High & Overall & Primary & Middle & High & Overall \\
\midrule
\multicolumn{9}{c}{MLLMs: Text + Image as Input} \\
\midrule
Gemini2-thinking&\textbf{60.79}&\underline{57.75}&\textbf{52.02}&\textbf{55.47}&\textbf{62.06}&\textbf{59.52}&\textbf{54.18}&\textbf{57.36}\\
Qwen2.5-VL-32B&\underline{59.03}&\textbf{58.49}&\underline{51.51}& \underline{55.42}&\underline{57.91}&\underline{56.50}&49.17&\underline{53.35}\\
Qwen2.5-VL-72B&54.08&54.72&47.39&51.39&55.27&53.83&44.79&49.93\\
Gemini2-flash&56.70&55.17&45.69&51.08&54.63&51.12&41.95&47.34\\
QVQ-72B-preview&54.99&51.99&45.9&49.54&47.42&48.15&44.17&46.31\\
InternVL2.5-MPO-78B&50.32&48.22&41.55&45.43&50.23&46.15&37.86&42.82\\
InternVL2.5-MPO-38B&46.97&44.02&36.31&40.85&46.70&41.84&33.46&38.54\\
InternVL2.5-78B&45.49&43.62&36.04&40.41&47.29&42.10&33.05&38.53\\
GPT-4o&45.56&37.42&30.39&35.02&48.28&37.44&29.80&35.00\\
InternVL2.5-38B&41.7&39.89&31.82&36.45&42.33&36.74&28.09&33.40\\
InternVL2.5-MPO-26B&36.05&37.15&29.37&33.58&37.34&33.59&24.6&29.92\\
Qwen2.5-VL-7B&40.40&44.97&34.32&39.82&38.67&30.90&20.89&27.16\\
LLaVA-OneVision-72B&33.68&34.59&28.01&31.57&32.70&30.10&22.88&27.11\\
Qwen2-VL-72B&40.17&37.92&29.74&34.48&32.88&29.32&20.48&25.71\\
InternVL2.5-MPO-8B&33.12&33.41&25.68&29.94&35.37&30.46&21.53&26.93\\
InternVL2.5-MPO-4B&34.88&32.49&24.92&29.33&33.76&27.91&20.39&25.09\\
InternVL2.5-26B&33.81&34.23&26.18&30.60&32.55&27.85&19.75&24.66\\
InternVL2.5-8B&28.90&30.53&23.63&27.31&29.89&25.88&17.93&22.69\\
InternVL2-76B&35.82&32.17&24.22&28.95&33.41&23.21&15.89&20.85\\
InternVL2.5-4B&33.14&30.55&22.48&27.18&29.62&23.43&16.07&20.70\\
MiniCPM-o-2.6&38.34&29.78&21.64&26.91&32.24&21.36&14.08&19.09\\
Qwen2.5-VL-3B&34.39	&36.46	&26.88	&31.99&25.54	&18.98&11.98	&16.45\\
InternVL2-40B&31.43&31.89&23.66&28.18&23.01&15.97&9.91&13.90\\
InternVL2-8B&27.95&28.96&21.93&25.73&22.37&14.64&9.34&12.97\\
Qwen2-VL-7B&31.22&28.47&21.71&25.70&18.74&14.62&9.21&12.58\\
LLaVA-OneVision-7B&24.89&26.72&20.87&23.94&16.70&13.18&9.64&11.92\\
\midrule
\multicolumn{9}{c}{LLMs: Text + Captions as Input} \\
\midrule
O3-mini&56.64&52.25&48.89&51.14&57.49&52.95&\underline{50.05}&52.06\\
Deepseek-v3&52.07&47.82&40.60&44.97&57.62&50.38&42.69&47.60\\
O1-mini&55.02&46.38&39.07&43.89&56.19&45.66&38.75&43.51\\
\bottomrule
\end{tabular}
}
\label{tab:main_table}
\vspace{-1.25pt}
\end{table}

(6) Hallucination Error: Factual errors, logical inconsistencies, or unwarranted inferences (e.g., answers contradicting known facts, logical contradictions, or baseless assumptions).

(7) Calculation  Error: Specific mistakes in mathematical operations or algebraic manipulations (e.g., arithmetic errors, incorrect equation solving, or flawed factorization).

(8) Incomplete Answer Error: Failing to provide a final answer or omitting parts of the answer (e.g., only addressing some sub-questions in a multi-part problem).
\subsection{Models for Solution Generation}\label{appendix:Models and Prompt for Solution Generation}
To more fully simulate the problem solving processes of MLLMs in real world scenarios, we employ the following 40 models to construct the problem Step-by-Step solution: Qwen2-VL(2B, 7B, 72B), Qwen2.5-VL(3B, 7B, 32B, 72B), GPT-4o, Gemini2-flash, Gemini2-thinking, O3-mini, O1-mini, InternVL2(4B, 8B, 26B, 40B, 76B), InternVL2.5(4B, 8B, 26B, 38B, 78B), QVQ-72B-preview, InternVL2.5-MPO(4B, 8B, 26B, 38B, 78B), LLaVA1.6(7B, 13B, 34B, 72B, 110B), and LLaVA-OneVison-(7B, 72B).
\subsection{Prompt for K12-PEM-800K Generation}\label{appendix:Prompt for K12-PEM-800K Generation}
\begin{figure}[!t]  
    \centering
    \includegraphics[width=1\textwidth]{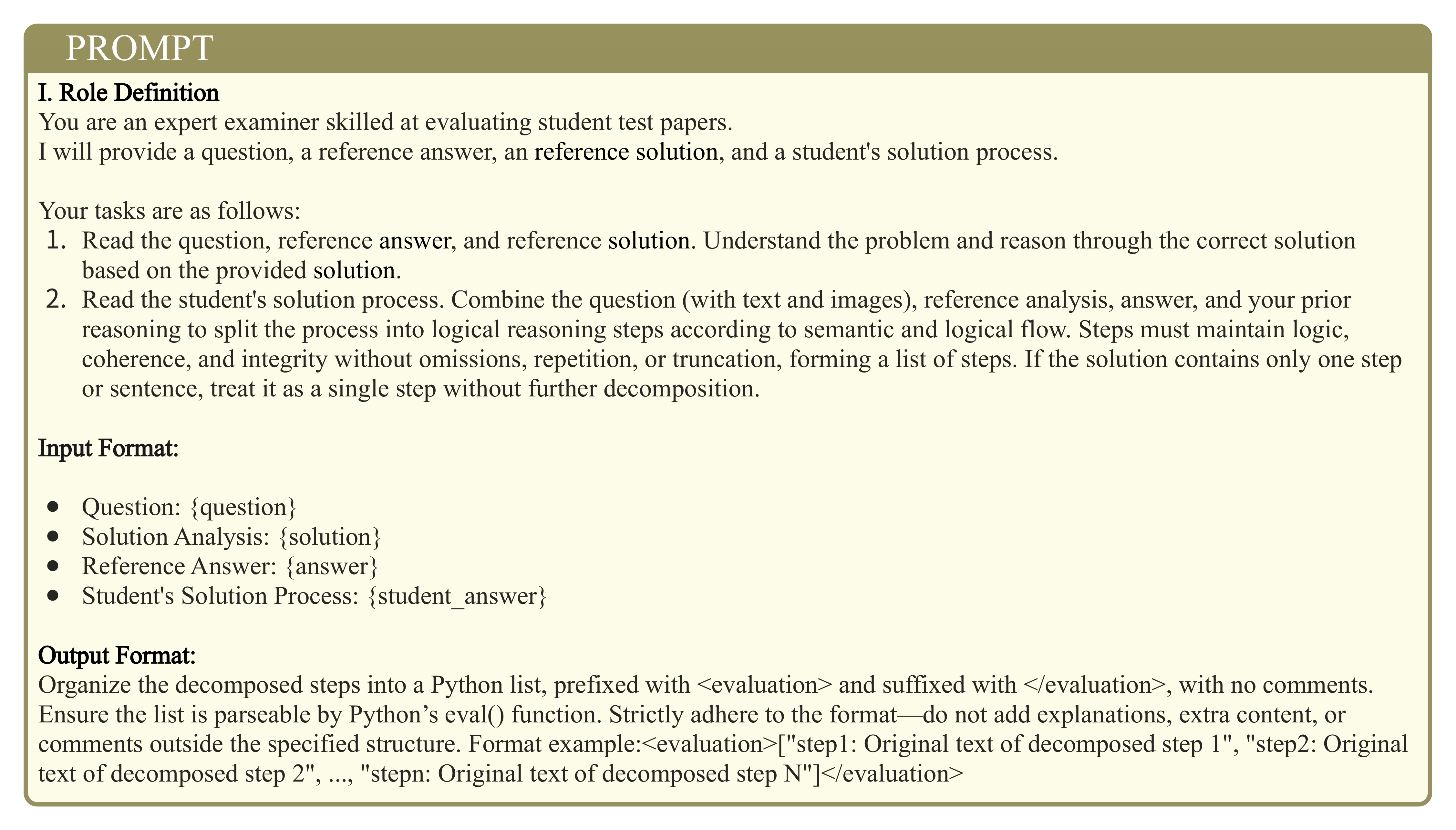}
    \caption{Prompt for split solution to steps. We have the GPT-4o decompose complete student responses into logically independent problem-solving steps for subsequent annotation. We provide their corresponding English translations.}
    \label{fig:appendix_K12-800k_split}
\end{figure}

\begin{figure}[!t]  
    \centering
    \includegraphics[width=1\textwidth]{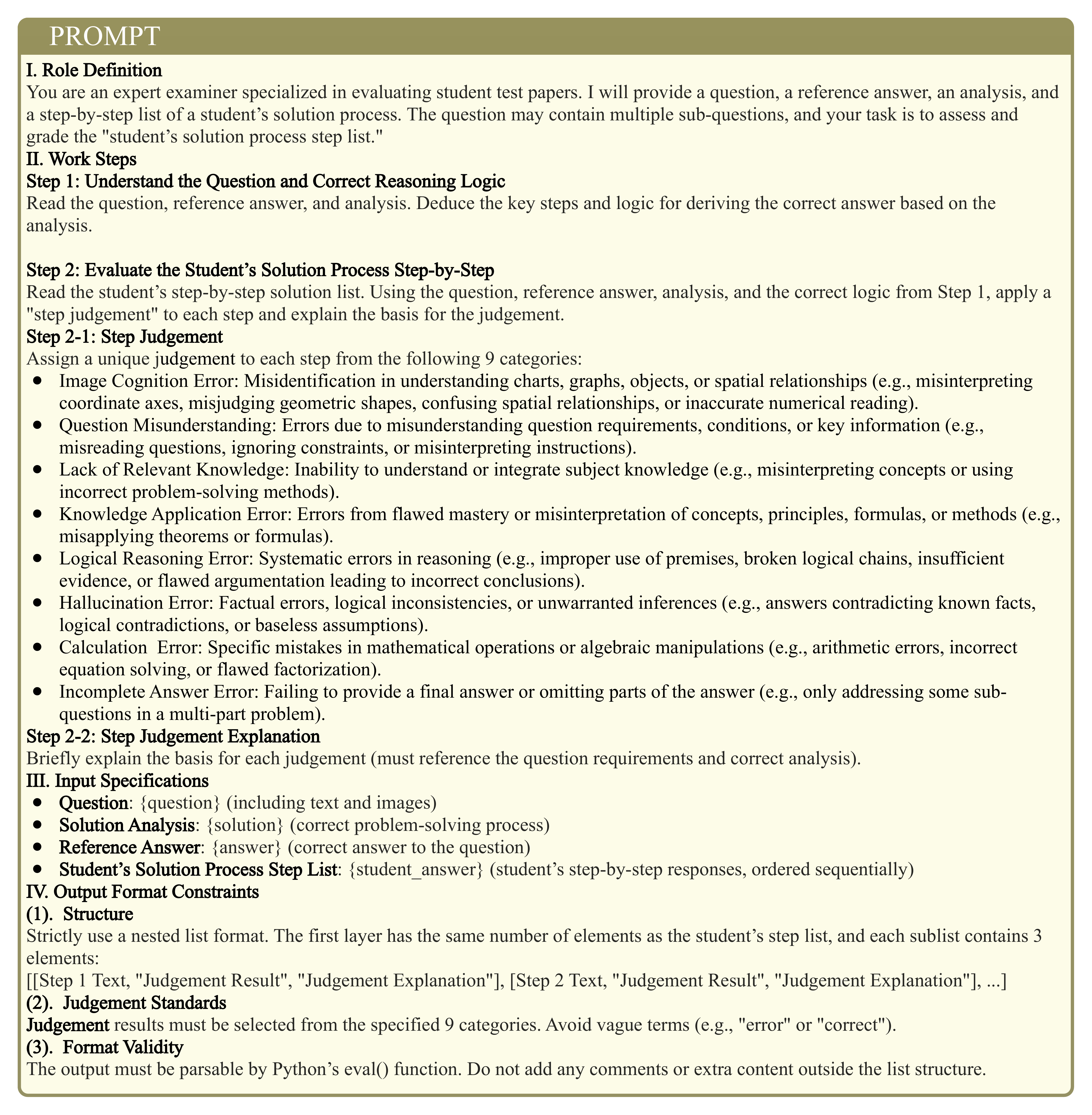}
    \caption{Prompt for step-wise judgement. We defined nine Step-wise labels, enabling the MLLM to judge each step with one of Step-wise labels according to the definitions. We provide their corresponding English translations.}
    \label{fig:appendix_K12-800k_judgement}
\end{figure}

The development of K12-PEM-800K primarily involves two stages: decomposing the MLLMs' solutions into step-by-step reasoning paths, judging and explaining each reasoning step. The corresponding prompts for each stage are illustrated in Figures \ref{fig:appendix_K12-800k_split} and \ref{fig:appendix_K12-800k_judgement}, respectively.
\subsection{Data Filtering}\label{appendix:Data Filtering}
We ensure data quality through rigorous data filtering processes, implementing four quality control mechanisms:
(1) Format Accuracy: Remove samples that deviate from the predefined format — each step in the student’s solution must be annotated as a tuple containing the step description, correctness, error type, and a brief explanation. Samples with mismatched step counts between annotations and student solutions are also discarded.
(2) Annotation Accuracy: Exclude samples with contradictory or incomplete annotations, such as steps marked as incorrect but lacking error type or cause descriptions.
(3) Question Coverage Assurance: Ensure that each question in K12-Vista appears at least three times in K12-PEM-800K.
(4) Error Type Balance: Maintain a balanced proportion of each error type to ensure diversity in the dataset.
\subsection{More Result about K12-Vista}\label{appendix:More Result about K12-Vista}
In this section, we present additional model results on K12-Vista, as shown in Table 6.
\section{K12-PEMBench Detail}
We have manually and carefully constructed a K2-PEMBench dataset containing 3,000 data points, covering diverse question types across 5 disciplines, to test models' capabilities in evaluating problem solving processes.
\subsection{Quality Assessment}\label{appendix:appendix_K12-PEMBench_Quality Assessment}
\begin{wraptable}[9]{r}{0.49\textwidth}
\vspace{-10pt}
\caption{The High-Quality Rate of 100 selected Samples} 
\centering
\resizebox{0.5\textwidth}{!}{%
\begin{tabular}{lcc} 
        \toprule 
        \textbf{Set} & \textbf{Question and Solution}& \textbf{Step-wise Label} \\
        \midrule 
        \textbf{Math} & 99.50  &93.00 \\
        \textbf{Physics} & 100.00  &95.50 \\
        \textbf{Chemistry} & 98.00  &88.50 \\
        \textbf{Biology} & 99.00  &87.00 \\
        \textbf{Geography} & 98.50  &86.00 \\
        \textbf{Overall} & 99.00  &90.00 \\
        \bottomrule 
    \end{tabular}}
    \label{tab:appendix_k12pembench_check} 
\end{wraptable}
We randomly sampled 100 questions from each subject, ensuring coverage of different grades. Three data reviewers verified the logical consistency of the questions with students' problem solving approaches, and the correctness of  step-wise label. The inspection results are presented in Table \ref{tab:appendix_k12pembench_check}, demonstrating high-quality rates of 99\% for questions and student solutions, 90\% for step-wise label.
\section{K12-PEM Train Detail}\label{appendix:K12-PEM Detail}
In the SFT phase, where the model directly evaluates the student's solution $x_S$ based on the problem $x_P^j$ and its final answer $x_A^j$ and reference solution $x_R^j$. For each solution, the model decomposes the student solution $x_S$ into steps and evaluates each step $s_i$, organizing them into a list where each element takes the form $[s_i, j_i, r_i]$ representing the step description, judgment, and explanation, respectively. The final structured output is formalized as: $[[s_i, j_i, r_i]_{i = 1}^M]$, $M$ denotes the total number of reasoning steps.

The training set for this task can be expressed as: $D = \{ [[s_i, j_i, r_i]_{i = 1}^M]\}_{j = 1}^N$, where $y^j$ represents the ground - truth step annotations and $N$ denotes the dataset size. During training, the model minimizes the cross - entropy loss between its predictions and the ground - truth annotations:
\begin{equation}
\begin{split}
\mathcal{L}(\theta, D) = -\frac{1}{N} \sum_{j = 1}^{N} \left[ \sum_{t = 1}^{|y^j|} \log p(y_t^j | x_P^j, x_A^j, x_S^j, x_R^j, y_{<t}^j; \theta) \right]
\end{split}
\end{equation}
where $y_{t}^j$ denotes the $t$-th token in the ground - truth sequence, $y_{<t}^j$ represents the preceding tokens.

We fine-tune Qwen2.5-VL-7B-Instruct on K12-PEM-800K. The training is conducted on 64 H200 GPUS. The global batch size is set to 128, with per-device batch size of 2 and gradient accumulation steps of 4. Additionally, we applied weight decay of 0.05 to regularize the training process and prevent overfitting. The models are trained with a learning rate of 2.0e-6, also using a cosine learning rate scheduler
and a warmup ratio of 0.1. Both fine-tuning processes utilize mixed-precision training (bf16) to accelerate computation and
reduce memory usage.
\section{Results}
\subsection{Cases Study}
\begin{table}[!t]
\caption{Accuracy of MLLMs across every step-wise labels. Overall performance represents the average accuracy across all step-wise labels. CS: correct step; ICE:mage Cognition Error; QM: Question Misunderstanding; LRK:Lack of Relevant Knowledge; KAE: Knowledge Application Error; LRE: Logical Reasoning Error; HE: Hallucination Error; CE: Calculation Error; IAE: Incomplete Answer Error;}
\centering
\resizebox{\textwidth}{!}{%
\begin{tabular}{l|cccccccccc} 
\toprule 
        \textbf{Model} & Overall&CS&ICE&QM&LRK&KAE&LRE&HE&CE&IAE \\
        \midrule 
        \textbf{K12-PEM} &69.38&94.33&53.55&34.51&24.87&44.13&30.14&25.32&48.87&68.69 \\
        \textbf{GPT-4o}&63.90&89.17&47.92&27.40&7.95&41.68&29.75&20.74&45.76&61.38 \\
        \textbf{InternVL2.5-MPO-78B} & 63.44&94.10&37.12&15.55&5.81&43.31&29.36&13.64&44.92&52.14 \\
        \textbf{Qwen2.5-VL-72B} & 63.17&89.83&43.54&29.53&3.28&41.27&22.97&17.51&45.20&62.21 \\
        \textbf{InternVL2.5-78B} & 60.38&92.62&25.76&17.01&6.82&39.53&25.29&12.85&33.33&39.45 \\
        \textbf{Qwen2-VL-72B } & 54.89&84.78&14.64&23.94&2.65&33.71&25.19&9.23&31.92&41.93 \\
        \textbf{InternVL2.5-MPO-8B } & 53.19&92.11&14.04&2.92&4.92&11.03&3.78&3.39&15.82&19.45 \\
        \textbf{Qwen2.5-VL-7B } & 48.74&81.88&3.74&4.80&19.44&11.95&27.13&4.73&10.45&22.34 \\
        \textbf{InternVL2.5-8B } & 46.96&72.67&17.77&14.76&2.65&24.92&28.10&11.04&26.27&21.66 \\
\bottomrule
\end{tabular}
}
\label{tab:appendix_K12_pemBench_result}
\vspace{-1.25pt}
\end{table}

\begin{figure}[!t]  
    \centering
    \includegraphics[width=1\textwidth]{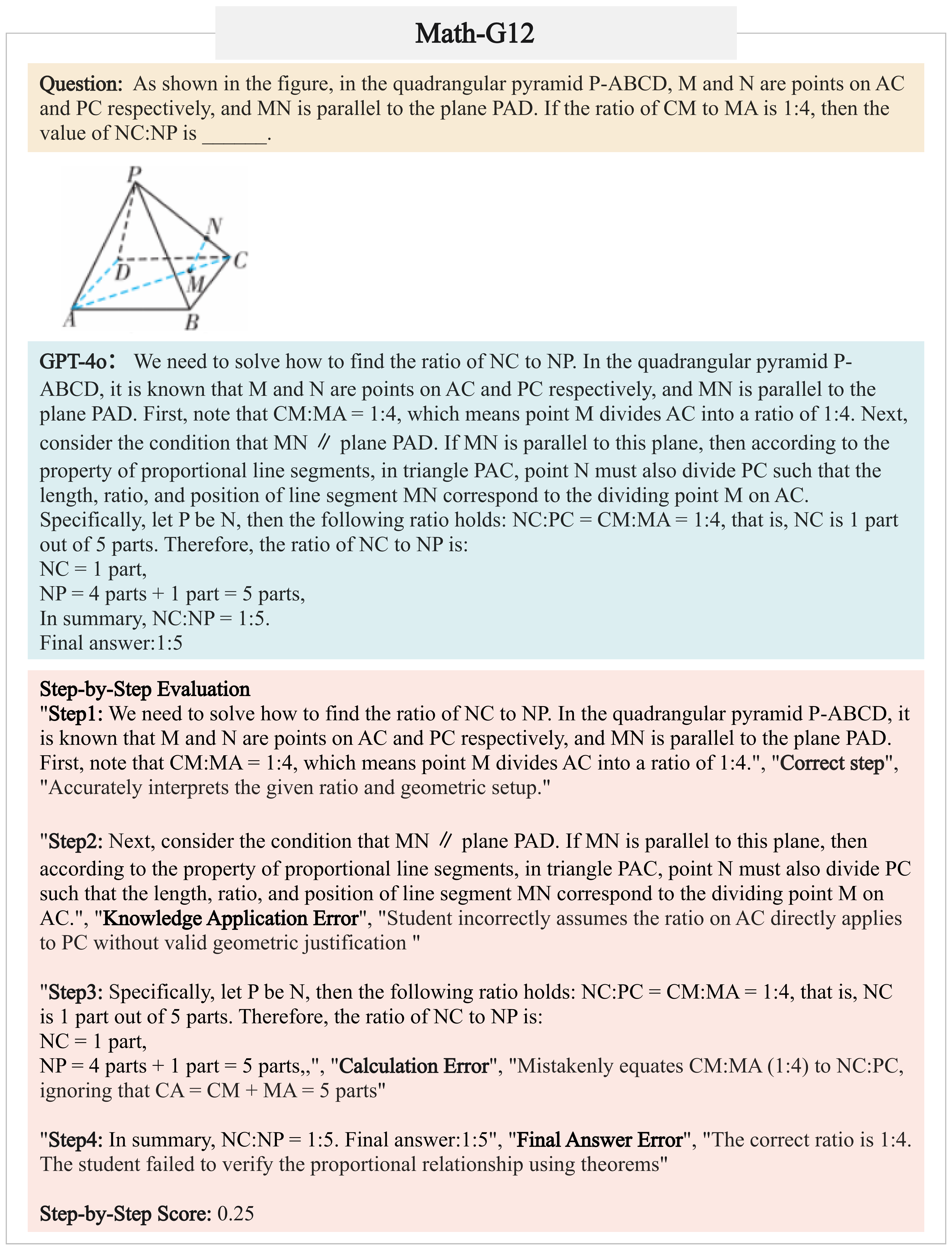}
    \caption{ A sample error case of Math from response of GPT-4o. We provide their corresponding English translations.}
    \label{fig:math_case}
\end{figure}

\begin{figure}[!t]  
    \centering
    \includegraphics[width=1\textwidth]{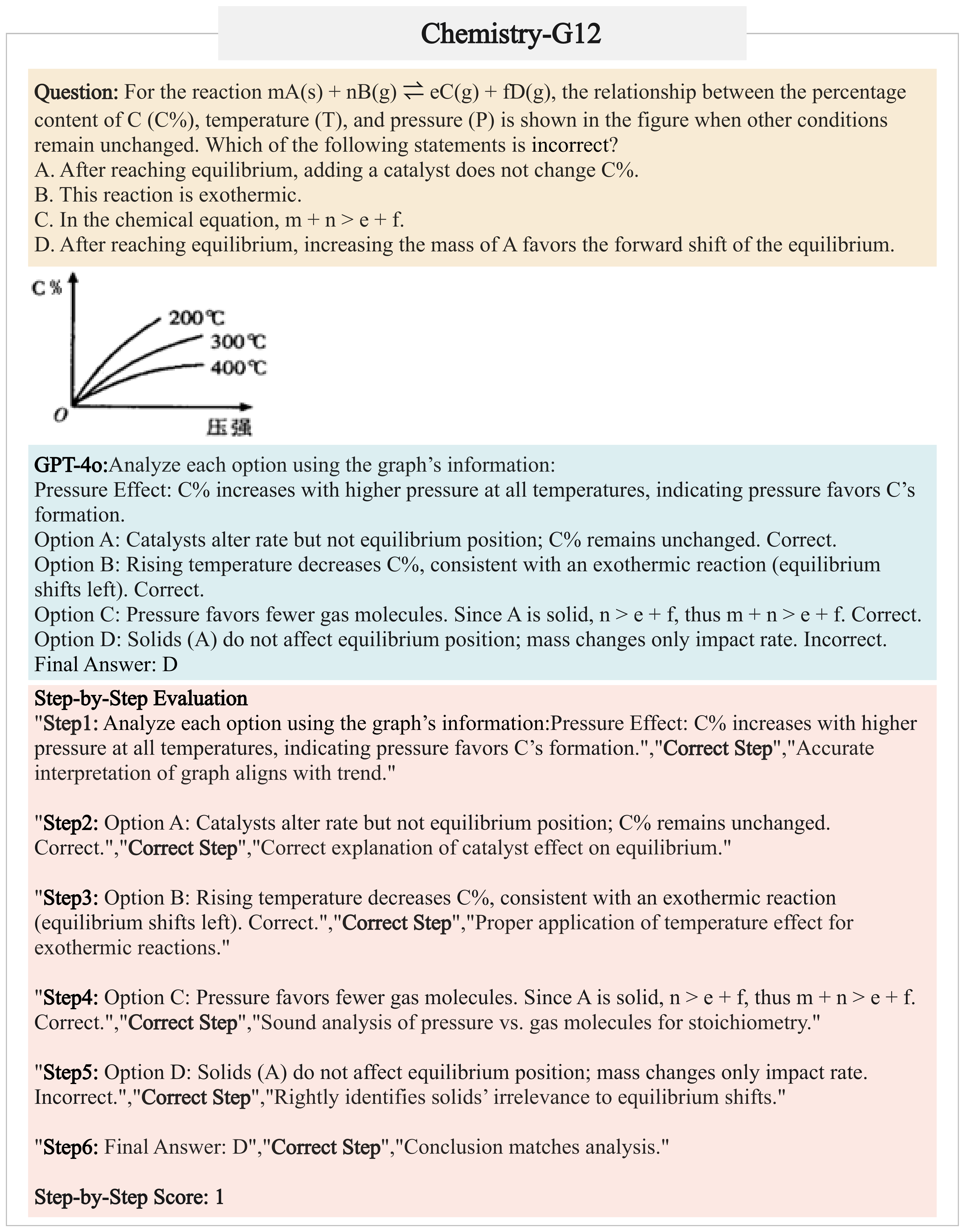}
    \caption{ A sample correct case of Chemistry from response of GPT-4o. We provide their corresponding English translations.}
    \label{fig:chemistry_case}
\end{figure}

\begin{figure}[!t]  
    \centering
    \includegraphics[width=1\textwidth]{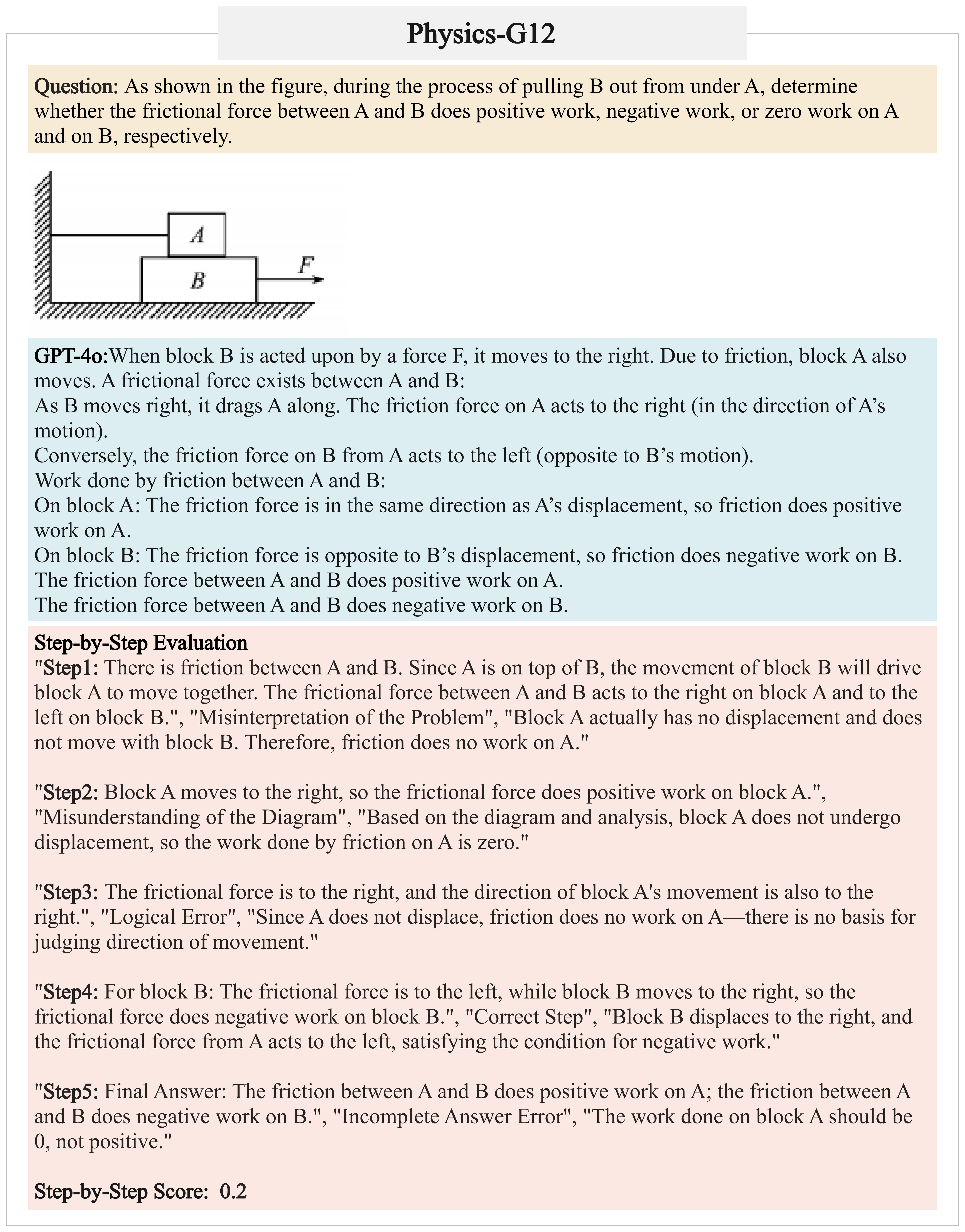}
    \caption{ A sample correct case of Phtsics from response of GPT-4o. We provide their corresponding English translations.}
    \label{fig:physics_case}
\end{figure}

We have selected one sample from three subjects in K12-Vista, as shown in Figures \ref{fig:math_case}, \ref{fig:chemistry_case}, \ref{fig:physics_case}.
\subsection{K12-PEMBench Result}\label{appendix:K12-PEMBench Result}
We evaluated the process evaluation capabilities of various models on K12-PEMBench, as detailed in Table \ref{tab:appendix_K12_pemBench_result}.


\end{document}